\documentclass{article}
\usepackage{ijcai18}

\usepackage{times}
\usepackage{xcolor}
\usepackage{soul}
\usepackage[utf8]{inputenc}
\usepackage[small]{caption}

\usepackage{pdfsync}

\usepackage{booktabs}       % professional-quality tables
\usepackage{amsfonts}       % blackboard math symbols
\usepackage{nicefrac}       % compact symbols for 1/2, etc.
\usepackage{microtype}      % microtypography
\usepackage{listings}

\usepackage{sidecap}
\usepackage{tabularx}
\usepackage{pbox}
\usepackage{csquotes}
\usepackage{booktabs}

\usepackage{graphicx} % more modern
\usepackage{subfigure}
\usepackage{caption}

\usepackage{amsmath, amsthm, amssymb}
\newtheorem{theorem}{Theorem}

\usepackage{algorithm}
\usepackage[noend]{algpseudocode}

\usepackage{enumitem}

\newcommand{\Actions}{\mathcal{A}}
\newcommand{\States}{\mathcal{S}}

\newcommand{\Fmat}{\mathbf{F}}
\newcommand{\Hmat}{\mathbf{H}}
\newcommand{\Kmat}{\mathbf{K}}

\newcommand{\defeq}{\mathrel{\overset{\makebox[0pt]{\mbox{\normalfont\tiny\sffamily def}}}{=}}}
\newcommand*{\argmin}{\mathop{\mathrm{argmin}}}

\newcommand{\Sigmamat}{\boldsymbol{\Sigma}}

\newcommand{\avec}{\mathbf{a}}
\newcommand{\bvec}{\mathbf{b}}
\newcommand{\cvec}{\mathbf{c}}

\newcommand{\svec}{\mathbf{s}}

\newcommand{\xvec}{\mathbf{x}}
\newcommand{\yvec}{\mathbf{y}}

\newcommand{\muvec}{\boldsymbol{\mu}}
\newcommand{\phivec}{\boldsymbol{\phi}}

\newcommand{\xdim}{d}

\newcommand{\nsamples}{T}

\newcommand{\stepsize}{\alpha}
\newcommand{\ksize}{b}
\newcommand{\nreplay}{m}

\newcommand{\inv}{{-1}}

\newcommand{\eye}{\mathbf{I}}

\newcommand{\E}{\mathbb{E}}

\newcommand{\citep}[1]{\cite{#1}}
\newcommand{\citet}[1]{\cite{#1}}

\newcommand{\myparagraph}[1]{\vspace{0.1cm}
\noindent \textbf{#1}}

\title{Organizing Experience: A Deeper Look at Replay Mechanisms for Sample-based Planning in Continuous State Domains}

\author{
Yangchen Pan \textsuperscript{1},
Muhammad Zaheer \textsuperscript{1},
Adam White \textsuperscript{1,2},
Andrew Patterson \textsuperscript{3},
Martha White \textsuperscript{1}
\\
\textsuperscript{1}University of Alberta, 
\textsuperscript{2}DeepMind, 
\textsuperscript{3}Indiana University\\
pan6@ualberta.ca,
mzaheer@ualberta.ca,
adamwhite@google.com,\\
andnpatt@iu.edu,
whitem@ualberta.ca
}

\begin{document}

\maketitle

%\twocolumn[
%\icmltitle{Revisiting Dyna for control in continuous-state domains}
%
%\begin{icmlauthorlist}
%\icmlauthor{Andrew Patterson}{iub}%{mkschlegel@umail.iu.edu}
%%\icmladdress{Department of Computer Science, Indiana University, Bloomington}
%\icmlauthor{Yangchen Pan}{iub}%{yangpan@umail.iu.edu}
%%\icmladdress{Department of Computer Science, Indiana University, Bloomington}
%\icmlauthor{Adam White}{iub}%{jiecchen@umail.iu.edu}
%%\icmladdress{Department of Computer Science, Indiana University, Bloomington}
%\icmlauthor{Martha White}{iub}%{martha@indiana.edu}
%\end{icmlauthorlist}
%% \icmladdress{Department of Computer Science, Indiana University, Bloomington}
%\icmlaffiliation{iub}{Department of Computer Science, Indiana University, Bloomington}
%\icmlcorrespondingauthor{Martha White}{martha@indiana.edu}
%
%\icmlkeywords{active set selection, submodular maximization, kernel representations}
%
%]

\begin{abstract}
Model-based strategies for control are critical to obtain sample efficient learning. Dyna is a planning paradigm that naturally interleaves learning and planning, by simulating one-step experience to update the action-value function. This elegant planning strategy has been mostly explored in the tabular setting. The aim of this paper is to revisit sample-based planning, in stochastic and continuous domains with learned models. We first highlight the flexibility afforded by a model over Experience Replay (ER). Replay-based methods can be seen as stochastic planning methods that repeatedly sample from a buffer of recent agent-environment interactions and perform updates to improve data efficiency. We show that a model, as opposed to a replay buffer, is particularly useful for specifying which states to sample from during planning, such as predecessor states that propagate information in reverse from a state more quickly. We introduce a semi-parametric model learning approach, called Reweighted Experience Models (REMs), that makes it simple to sample next states or predecessors. We demonstrate that REM-Dyna exhibits similar advantages over replay-based methods in learning in continuous state problems, and that the performance gap grows when moving to stochastic domains, of increasing size.
%Model-based strategies for control are critical to obtain sample efficient learning. Dyna is a planning paradigm that naturally interleaves learning and planning, by simulating one-step experience to update the action-value function. This elegant planning strategy has largely only been explored for the tabular setting. The aim of this paper is to revisit sample-based planning, in stochastic and continuous domains with learned models. We first highlight the flexibly afforded by a model over Experience Replay (ER). Replay based methods can be seen as stochastic planning methods that repeatedly sample from a buffer of recent agent-environment interactions and perform updates to improve data efficiency. We show that a model, as opposed to a replay buffer, is particularly useful for specifying which states to sample during planning through the use of predecessor states that propagate information in reverse from a state more quickly and through the generation of on-policy transitions. We introduce a lightweight model learning approach, called Reweight Experience Models (REMs), which naturally enable either sampling next states or predecessor states. We demonstrate that REM-Dyna exhibits similar advantages over replay-based methods in learned in continuous state problems, and that the performance gap grows when moving to stochastic domains, of increasing size.
\end{abstract}

\section{Introduction}
Experience replay has become nearly ubiquitous in modern large-scale, deep reinforcement learning systems \cite{schaul2016prioritized}. The basic idea is to store an incomplete history of previous agent-environment interactions in a {\em transition buffer}. During planning, the agent selects a transition from the buffer and updates the value function as if the samples were generated online---the agent {\em replays} the transition. There are many potential benefits of this approach, including stabilizing potentially divergent non-linear Q-learning updates, and mimicking the effect of multi-step updates as in eligibility traces.  %TODO: im replacing "experience" with "sample" everywhere

Experience replay (ER) is like a model-based RL system, where the transition buffer acts as a model of the world \cite{lin1992self}. Using the data as your model avoids model errors that can cause bias in the updates (c.f. \citep{bagnell2001autonomous}).
%, but causes updating to occur off-policy because the samples in the replay buffer where sampled according to many different polices (different from the agents current one). Due to the explosion of recent progress, in sound and efficient off-policy learning algorithms, this appears to be a wise tradeoff.
One of ER's most distinctive attributes as a model-based planning method, is that it does not perform multistep rollouts of hypothetical trajectories according to a model; rather previous agent-environment transitions are replayed randomly or with priority from the transition buffer. Trajectory sampling approaches such as PILCO \citep{deisenroth2011pilco},
Hallucinated-Dagger \citep{talvitie2017self}, and
CPSRs \citep{hamilton2014efficient}, unlike ER, can rollout unlikely trajectories ending up in hypothetical states that do not match any real state in the world when the model is wrong \citep{talvitie2017self}. ER's stochastic one-step planning approach was later adopted by Sutton's Dyna architecture \citet{sutton1991integrated}.

Despite the similarities between Dyna and ER,
%Experience replay, has been used in many successful demonstrations of deep Q-learning, however,
there have been no comprehensive, direct empirical comparisons comparing the two and their underlying design-decisions.
%, many of the design decisions underlying each approach have not be thoroughly investigated in the function approximation setting.
ER maintains a buffer of transitions for replay, and Dyna a search-control queue composed of stored states and actions from which to sample.
%\footnote{Comparison between Dyna and model free methods have been investigated \citep{adam2012experience}. Further, \citet{gu2016continuous} showed a complementary comparison between a model-based approach and ER, by using a model to generate multi-step rollouts to for addition to the replay buffer.}  %TODO i dropped "---especially with function approximation---" from the previous sentence because it was never explained.
There are many possibilities for how to add, remove and select samples from either ER's transition buffer or Dyna's search-control queue.
% to and replay from ER's transition buffer---let alone deleting older transitions from the buffer.
It is not hard to imagine situations where a Dyna-style approach could be better than ER.
For example, because Dyna models the environment, states leading into high priority states---predecessors---can be added to the queue, unlike ER.
Additionally, Dyna can choose to simulate on-policy samples, whereas ER can only replay (likely off-policy) samples previously stored.
%For example, in situations where states are often revisited the transition buffer could fill with redundant transitions, unlike a model.
In non-stationary problems, small changes can be quickly recognized and corrected in the model. On the other hand, these small changes might result in wholesale changes to the policy, potentially invalidating many transitions in ER's buffer.
%TODO: removed because we no longer study it: In addition if the environment changes in some way, a Dyna-style approach could potentially update the  parts of the model corresponding to the change, and then the next planning step could correct the value function. Experience replay, on the other hand, could suffer in these situations because the buffer could become polluted with transitions that are no longer valid due to the change in the environment.
%
% TODO: since we don't test this, I'm not sure if we should say this
%Finally, most variants of ER slowly remove transition from the buffer via recency. Those transitions might correspond to important situations, where a model by design should remember these situations.
%In summary, a model-based approach such as Dyna, has the potential to be much more compact and data efficient than experience replay.
It remains to be seen if these differences manifest empirically, or if the additional complexity of Dyna is worthwhile.
% the these natural extensions of Dyna and ER is worthy compared to the simplicity of conventional recency-based ER.

In this paper, we develop a novel semi-parametric Dyna algorithm, called REM-Dyna, that provides some of the benefits of both Dyna-style planning and ER.
We highlight criteria for learned models used within Dyna, and propose Reweighted Experience Models (REMs) that are data-efficient, efficient to sample and can be learned incrementally.
We investigate the properties of both ER and REM-Dyna, and highlight cases where ER can fail, but REM-Dyna is robust.
%specifically focusing on how to select which transitions should be added to the buffer for ER and which prototypes to add to REM's model. We highlight cases where ER can fail, but REM-Dyna is robust.
%Further, we demonstrate that REM-Dyna retains it's efficiency advantages against model-free methods with small models (set of prototype
%transitions), compared with ER which often requires a large buffer to achieve competitive performance. %TODO is this true?
Specifically, this paper contributes both (1) a new method extending Dyna to continuous-state domains---significantly outperforming previous attempts \cite{sutton2008dyna}, and (2) a comprehensive investigation of the design decisions critical to the performance of one-step, sample-based planning methods for reinforcement learning with function approximation. An \textbf{Appendix is publicly available on arXiv}, with theorem proof and additional algorithm and experimental details.

\section{Background}
We formalize an agent's interaction with its environment as a discrete time Markov Decision Process (MDP).
% Adam: We in fact do not rely on this, so I'm omitting it.
%\footnote{Our experiments feature infinite MDPs, however, for simplicity we assume a large finite state-action space.}
On each time step $t$, the agent observes the state of the MDP $S_t \in\mathcal{S}$, and selects an action $A_t \in\mathcal{A}$, causing a transition to a new state $S_{t+1} \in\mathcal{S}$ and producing a scalar reward on the transition $R_{t+1}\in\mathbb{R}$. The agent's objective is to find an optimal policy $\pi:\mathcal{S}\times\mathcal{A}\rightarrow[0,1]$, which maximizes the expected return $Q_\pi(\svec,a)$ for all $\svec, a$, where
%$G_t = \sum_{k=0}^\infty \gamma^{k} R_{t+k+1}$
$G_t \defeq R_{t+1} + \gamma(S_t, A_t, S_{t+1}) G_{t+1}$,
$\gamma: \States \times \Actions \times \States \in[0,1]$,
and $Q_\pi(s,a) = \mathbb{E}[G_t| S_t=s, A_t=a; A_{t+1:\infty}\sim\pi]$, with future states and rewards are sampled according to the one-step dynamics of the MDP.
%$Q_\pi: \mathcal{S}\times\mathcal{A}\rightarrow\mathbb{R}$:
%\begin{align*}
%\pi^\star = \max_\pi Q_\pi(s,a), \text{for all}~s\in \mathcal{S}~\text{and}~a\in\mathcal{A}
%\end{align*}
%where
%$Q_\pi(s,a) = \mathbb{E}[G_t| S_t=s, A_t=a; A_{t+1:\infty}\sim\pi]$, $\gamma: \States \times \Actions \times \States \in[0,1]$, and future states and rewards are sampled according to the one-step dynamics of the MDP.
The generalization to a discount function $\gamma$ allows for a unified specification of episodic and continuing tasks \citep{white2017unifying}, both of which are considered in this work.
%: ${\bf pr}(s',r|s,a)$.

In this paper we are concerned with model-based approaches to finding optimal policies. In all approaches we consider here the agent forms an estimate of the value function from data: $\hat{q}_\pi(S_t,A_t,\theta) \approx \mathbb{E}[G_t| S_t=s, A_t=a]$. The value function is parameterized by $\theta\in\mathbb{R}^n$ allowing both linear and non-linear approximations.
We consider sample models, that given an input state and action need only output one possible next state and reward, sampled according to the one-step dynamics of the MDP: $M: \mathcal{S} \times \mathcal{A} \rightarrow \mathcal{S}\times \mathbb{R}$.
% ADAM, TODO: We don't every really talk about probability distribution models anymore, and the distinction
%We consider both stochastic one-step {\em sample-based models} and {\em probability distribution models}. A distribution model captures the one-step transition dynamics of the MDP, specifying the full distribution over next states and rewards, given an input state and action: $M_d: \mathcal{S} \times \mathcal{A} \times \mathcal{S} \times \mathbb{R} \rightarrow [0,1]$.
%A sample model, on the other hand, given an input state and action need only output one possible next state and reward, sampled according to the one-step dynamics of the MDP: $M_s: \mathcal{S} \times \mathcal{A} \rightarrow \mathcal{S}\times \mathbb{R}$.
%In this paper we investigate a family of methods that use (one-step) transitions sampled from a model, to update an action-value function.

%Both types of models can be used to to simulate hypothetical trajectories of future states, actions, and, rewards, however
In this paper, we focus on stochastic one-step planning methods, where one-step transitions are sampled from a model to update an action-value function. The agent interacts with the environment on each time step, selecting actions according to its current policy (e.g., $\epsilon$-greedy with respect to $\hat{q}_\pi$), observing next states and rewards, and updating $\hat{q}_\pi$. Additionally, the agent also updates a model with these {\em observed sample} transitions $<S_t,A_t,S_{t+1},R_t>$ on each time step. After updating the value function and the model, the agent executes $\nreplay$ steps of planning. On each planning step, the agent samples a start state $S$ and action $A$ in some way (called {\em search control}), then uses the model to {\em simulate} the next state and reward. Using this {\em hypothetical transition} the agent updates $\hat{q}_\pi$ in the usual way. In this generic framework, the agent can interleave learning, planning, and acting---all in realtime. Two well-known implementations of this framework are ER \citep{lin1992self}, and the Dyna architecture \citep{sutton1991integrated}.

% MARTHAC: I sort of like this paragraph,TODO
%There are many details regarding how the model is represented and updated, and how the planning step should be implemented. All these decisions can dramatically effect the agent's overall performance, and that is the topic of the remainder of this paper.

%\section{Planning approaches for updating the action-value function}
\section{One-step Sample-based Planning Choices}
There are subtle design choices in the construction of stochastic, one-step, sample-based planning methods that can significantly impact performance.
These include how to add states and actions to the search-control queue for Dyna, how to select states and actions from the queue, and how to sample next states. These choices influence the design of our REM algorithm, and so we discuss them in this section.

One important choice for Dyna-style methods is whether to sample a next state, or compute an expected update over all possible transitions.
%%For each update to $Q(\svec, a)$, one can either use the full dynamic programming update or a sample update.
%%The full dynamic programming update, for the tabular setting, is
%%$Q(\svec, a) = \sum_{\svec', r, a'} P(\svec', r) \pi(\svec', a') (r + \gamma(\svec,a,\svec') Q(\svec',a'))$;
%Under function approximation, the full dynamic programming update corresponds to estimating the expected next feature vector
%or expected update to the value-function parameters.
A sample-based planner samples $\svec', r, \gamma$, given $\svec,a$, and stochastically updates $\hat{q}_\pi(\svec, a,\theta)$.
An alternative is to approximate full dynamic programming updates, to give an expected update, as done by
stochastic factorization approaches \citep{barreto2011reinforcement,kveton2012kernel,barreto2014policy,yao2014pseudo,barreto2016incremental,pires2016policy}, kernel-based RL (KBRL) \citet{ormoneit2002kernel}, and kernel mean embeddings (KME) for RL \citep{grunewalder2012modelling,vanhoof2015learning,lever2016compressed}.
Linear Dyna \citep{sutton2008dyna} computes an expected next reward and expected next feature vector for the update, which  corresponds to an expected update when $\hat{q}_\pi$ is a linear function of features.
We advocate for a sampled update, because approximate dynamic programming updates, such as KME and KBRL, are typically too expensive, couple the model and value function parameterization and are designed for a batch setting.
Computation can be more effectively used by sampling transitions. % and providing more lightweight updates.

There are many possible refinements to the search-control mechanism, including prioritization and backwards-search.
For tabular domains, it is feasible to simply store all possible states and actions, from which to simulate.
In continuous domains, however, care must be taken to order and delete stored samples.
A basic strategy is to simply store recent transitions $(\svec,a,\svec',r,\gamma)$ for the transition buffer in ER, or state and actions $(\svec, a)$ for the search-control queue in Dyna. This, however, provides little information about which samples would be most beneficial for learning.
Prioritizing how samples are drawn, based on absolute TD-error $|R + \gamma\max_a' \hat{q}_\pi(\svec', a') - \hat{q}_\pi(\svec, a)|$, has been shown to be useful for both tabular Dyna \citep{sutton1998reinforcement}, and ER with function approximation \citep{schaul2016prioritized}. When the buffer or search-control queue gets too large, one then must also decide whether to delete transitions based on recency or priority. In the experiments, we explore this question about the efficacy of recency versus priorities for adding and deleting.

%Though prioritization can be added to both,
ER is limited in using alternative criteria for search-control, such as backward search. A model allows more flexibility in obtaining useful states and action to add to the search-control queue. For example, a model can be learned to simulate {\em predecessor states}---states leading into (a high-priority) $\svec$ for a given action $a$. Predecessor states can be added to the search-control queue during planning, facilitating a type of backward search.
%This idea combines well with prioritizing the contents of the search-control queue based on TD-error for each transition. Indeed this
The idea of backward search and prioritization were introduced together for tabular Dyna \cite{peng1993efficient,moore1993prioritized}.  Backward search can only be applied in ER in a limited way because its  buffer is unlikely to contain transitions from multiple predecessor states to the current state in planning. \citet{schaul2016prioritized} proposed a simple heuristic to approximate prioritization with predecessors, by updating the priority of the most recent transition on the transition buffer to be at least as large as the transition that came directly after it. This heuristic, however, does not allow a systematic backward-search.
%Therefore, an additional design decision relies on the interaction between predecessor states, and priorities versus recency.

%design decision is how the addition of priorities, and deleting based on recency or priorities, interacts with the efficacy of predecessors states.
%These design choices for addition and deletion equally apply to REM's model, and the decision of how to delete old samples is likely to interact with the efficacy of predecessor states.

%In continuous domains, it is not possible to record all the agent-environment interactions in an infinite queue or buffer. Care must be taken to order the queue and delete transitions. Transitions $(\svec,a,\svec',r,\gamma)$, can be stored in ER's buffer according to recency or the priority of the transition---using the magnitude of the TD update most recently performed with the transition: $R + \gamma\max_a' \hat{q}_\pi(\svec', a') - \hat{q}_\pi(\svec, a)$. When the buffer gets too large, one could delete transitions based on recency or priority. As we show in our experiments, in some problems recency is a better choice. These design choices equally apply to REM's model, and in addition the decision of how to delete old samples interacts with the efficacy of predecessor states.

A final possibility we consider is using the current policy to select the actions during search control. Conventionally, Dyna draws the action from the search-control queue using the same mechanism used to sample the state. Alternatively, we can sample the state via priority or recency, and then query the model using the action the learned policy would select in the current state: $\svec', R, \gamma \sim \hat P(\svec, \pi(\svec), \cdot, \cdot, \cdot)$. This approach has the advantage that planning focuses on actions that the agent currently estimates to be the best. In the tabular setting, this {\em on-policy sampling} can result in dramatic efficiency improvements for Dyna \citep{sutton1998reinforcement}, while \citep{gu2016continuous} report improvement from on-policy sample of transitions, in a setting with multi-step rollouts. ER cannot emulate on-policy search control because it replays full transitions $(\svec,a,\svec',r,\gamma)$, and cannot query for an alternative transition if a different action than $a$ is taken.

\section{Reweighted Experience Models for Dyna}

%There are range of possible options for the sampling model within Dyna.
In this section, we highlight criteria for selecting amongst the variety of available sampling models,
 and then propose a semi-parametric model---called Reweighted Experience Models---as one suitable model that satisfies these criteria.

 \subsection{Generative Models for Dyna}\label{sec_generative}

Generative models are a fundamental tool in machine learning, providing a wealth of possible model choices.
We begin by specifying our desiderata for online sample-based planning and acting.
%For use within the Dyna formalism, however, there are additional criteria for them to be useful.
First, the model learning should be \emph{incremental and adaptive}, because the agent incrementally interleaves learning and planning.
Second, the models should be \emph{data-efficient}, in order to achieve the primary goal of improving data-efficiency of learning value functions.
%They can be learned more slowly, but still need to be themselves data-efficient and adaptive to be worthwhile within Dyna for the purposes of improving data-efficiency.
Third, due to policy non-stationarity, the models need to be \emph{robust to forgetting}:
% because the policy is changing, the distribution over states is constantly changing;
if the agent stays in a part of the world for quite some time, the learning algorithm should not overwrite---or forget---the model in other parts of the world.
Fourth, the models need to be able to be queried as \emph{conditional models}.
Fifth, \emph{sampling should be computationally efficient}, since a slow sampler will reduce the feasible number of planning steps.

% MARTHA to ADAM: we just don't have space. does anyone want this paragraph?
%We would like to emphasize that we do not disparage any of the generative models described below. Much of the purpose of this work
%is to revisit Dyna, relative to ER, for stochastic domains with learned models. Many models could be suitable, and any issues
%highlighted below could of course be overcome with new developments. Rather, below we highlight our reasoning for why we pursue the
%semi-parametric approach described in the next section.

Density models are typically learned as a mixture of simpler functions or distributions.
In the most basic case, a simple distributional form can be used, such as a Gaussian distribution for continuous random variables, or a categorical distribution for discrete random variables. For conditional distributions, $p(\svec' | \svec, a)$, the parameters to these distributions, like the mean and variance of $\svec'$, can be learned as a (complex) function of $\svec, a$. More general distributions can be learned using mixtures, such as mixture models or belief networks. A Conditional Gaussian Mixture Model, for example, could represent $p(\svec' | \svec, a) = \sum_{i=1}^\ksize \alpha_i(\svec, a) \mathcal{N}(\svec' | \muvec(\svec, a), \Sigmamat(\svec, a))$, where $\alpha_i, \muvec$ and $\Sigmamat$ are (learned) functions of $\svec, a$.
In belief networks---such as Boltzmann distributions---the distribution is similarly represented as a sum over hidden variables, but for more general functional forms over the random variables---such as energy functions.
%with $p(\svec', \svec, a) \propto \sum_{i=1}^\ksize \exp(-\text{Energy}((\svec', \svec, a), i))$.
To condition on $\svec, a$, those variables in the network are fixed both for learning and sampling.

%To learn a density or sampling model, typically a distributional form needs to be selected.
%For example, one can assume that the distribution is a Gaussian distribution for continuous random variables, or a categorical distribution for discrete random variables. The goal then is to learn the parameters of this distribution. Given these parameters, one then also
%has to be able to sample from that model. The choice of model, therefore, is informed both by ability to learn the parameters and the ability to sample. This typically restricts many approaches to learning the parameters of simpler distributions, like Gaussians, though the function class itself can be complex---such as using a deep neural network to learn the mean and covariance of observations on the next step, given the observation and action on this step.

Kernel density estimators (KDE) are similar to mixture models, but are non-parametric: means in the mixture are the training data, with a uniform weighting: $\alpha_i = 1/\nsamples$ for $\nsamples$ samples. KDE and conditional KDE is consistent \citep{holmes2007fast}---since the model is a weighting over observed data---providing low model-bias. Further, it is data-efficient, easily enables conditional distributions, and is well-understood theoretically and empirically. Unfortunately, it scales linearly in the data, which is not compatible with online reinforcement learning problems. Mixture models, on the other hand, learn a compact mixture and could scale, but are expensive to train incrementally and have issues with local minima.

%Mixture models and kernel density estimators use a weighted average of simpler distributions---such as Gaussians---to get a more complex distribution.
%These approaches are attractive, in that they can model many distributions and have well-understood theoretical and empirical properties \citep{}.
%Kernel density estimators are typically considered non-parametric, since the given data is used as centers for the kernels, the parameters for the Gaussians are not learned and there is a uniform weighting given to each mixture component. Mixture models are parametric, in that the centers and coefficients are learned. For incremental estimation, however, it is not straightforward to simply apply either model. Kernel density estimators are consistent---since they are simply a weighting over observed data---providing low model-bias; however, they grow with increasing data and so cannot be directly used for Dyna. Mixture models, on the other hand, are expensive to learn incrementally \citep{} and would be likely to suffer from forgetting, with centers being constantly adjusted to be best suited for most recent data.

Neural network models are another option, such as Generative Adversarial Networks \citep{goodfellow2014generative} and Stochastic Neural Networks \citep{sohn2015learning,alain2016gsn}.
Many of the energy-based models, however, such as Boltzmann distributions, require computationally expensive sampling strategies  \citep{alain2016gsn}.
%---such as Markov chain Monte Carlo \citep{alain2016gsn}. %Generative Adversarial Networks and
Other networks---such as Variational Auto-encoders---sample inputs from a given distribution, to enable the network to sample outputs.
%---including even simpler models such as Denoising Auto-encoders \citep{alain2016gsn}.
These neural network models, however, have issues with forgetting \citep{mccloskey1989catastrophic,french1999catastrophic,goodfellow2013empirical}, and require
more intensive training strategies----often requiring experience replay themselves.
%They are a promising to explore in future work,
%but it is outside the scope of this work to ameliorate both incremental training and forgetting issues.
%%Other networks, like Denoising Auto-encoders or Boltzmann distributions, can also be used to generate samples, but require more complex sampling
%%strategies \citep{alain2016gsn} like Markov chain Monte Carlo.

% TODO: ensure the below message is incorporated above
%There are several potential approaches to using non-parametric density estimators, within Dyna.
%Two recently popular non-parametric methods within reinforcement learning are kernel mean embeddings (KME) \citep{grunewalder2012modelling}
%and locally-weighted sample averages, such as in KBRL \citep{ormoneit2002kernel}. There are some difficulties, however, in using these approaches
%within Dyna, which requires smoothly interleaving planning and learning.
%KME and KBRL both couple the model parametrization
%and the value function representation; consequently, when the model changes, the value function needs to be recomputed.
%Further, both KME and KBRL are designed for batches of data, staging improvements to the action-value function and gathering
%new batches of data to enable stationary samples to facilitate estimation of the value function.
%This is problematic for incrementally interleaving planning and learning.
%% so we pursue a kernel density estimation approach as a more suitable direction.

 \subsection{Reweighted Experience Models}

%However, for completeness, we include these other non-parametric estimators in the appendix, and discuss the potential
%benefits they could have over REM-Dyna.

We propose a semi-parametric model to take advantage of the properties of KDE
and still scale with increasing experience.
The key properties of REM models are that 1) it is straightforward to specify and sample both forward and reverse models for predecessors---$p(\svec' | \svec, a)$ and $p(\svec | \svec', a)$---using essentially the same model (the same prototypes); 2) they are data-efficient, requiring few parameters to be learned; and 3) they can provide sufficient model complexity, by allowing for a variety of kernels or metrics defining similarity. %MARTHA: can the reader understand why we have both: $p(\svec' | \svec, a)$ and $p(\svec | \svec', a)$, ADAM: Hopefully by now we have specified that we care about predecessors, i added a bit more info

%
%In particular, because KDE
%Nonetheless, the power, simplicity and potentially low model-bias makes this general class of generative models attractive; below, we will develop a semi-parametric model, called Reweighted Experience Models, as an effective middle-ground between the two for use within Dyna.
%In particular, as we will show, the model is the most useful within Dyna
%when being able to query both next states, conditioned on a state and action, and query predecessors states, i.e., obtain a sample of a state that leads to the given state. The ability to switch between different conditioning is one of the biggest reasons we will select a non-parametric density estimator in the next section.

REM models consist of a subset of prototype transitions $\{ (\svec_i, a_i, \svec'_i, r_i, \gamma_i) \}_{i=1}^\ksize$, chosen from all $\nsamples$ transitions experienced by the agent, and their corresponding weights $\{c_i\}_{i=1}^\ksize$. These prototypes are chosen to be representative of the transitions, based on a similarity given by a \emph{product kernel} $k$
\small
\begin{align}
&p(\svec, a, \svec', r, \gamma | \svec_i, a_i, \svec_i', r_i, \gamma_i) \defeq k((\svec, a, \svec', r, \gamma), (\svec_i, a_i, \svec_i', r_i, \gamma_i)) \nonumber\\
&\hspace{1.0cm}\defeq k_\svec(\svec,\! \svec_i) k_a(a, a_i) k_{\svec', r, \gamma} ((\svec', r, \gamma), (\svec_i', r_i, \gamma_i)). \label{eq_product}
\end{align}
\normalsize
A product kernel is a product of separate kernels. It is still a valid kernel, but simplifies dependences and simplifies computing conditional densities, which are key for Dyna, both for forward and predecessor models. They are also key for obtaining a consistent estimate of the $\{c_i\}_{i=1}^\ksize$, described below.
%The model-structure imposed by REMs, with similarities to prototypes, combined with neural networks to learn embeddings or metrics.
%\footnote{To satisfy integrating to one, it is standard to use a normalized kernel: $k_{\svec}(\cdot, \svec_i)/ \int k_{\svec}(\svec, \svec_i) d \svec$.}
%

We first consider Gaussian kernels for simplicity. For states, %the Gaussian kernel is
\begin{equation*}
 k_\svec(\svec, \svec_i) = (2\pi)^{-\xdim/2} | \Hmat_s |^{-1/2} \exp(- (\svec - \svec_i)^\top \Hmat_s^\inv (\svec - \svec_i) )
\end{equation*}
with covariance $\Hmat_s$.
%which could simply be fixed to a diagonal covariance for some covariance $\sigma^2$.
%where the covariance could be a fixed diagonal or a sample covariance matrix scaled by the number of points in the model, $\Hmat_s = \ksize^\inv \Sigma_s$.
For discrete actions, the similarity is an indicator $k_a(a, a_i) = 1$ if $a = a_i$ and otherwise $0$.
For next state, reward and discount, a Gaussian kernel is used for $k_{\svec', r, \gamma}$ %
%\small
%\begin{align*}
%&k_{\svec', r} ((\svec', r, \gamma), (\svec_i', r_i, \gamma_i)) = (2\pi)^{-(\xdim+2)/2} | \Hmat_{\svec', r,\gamma} |^{-1/2}\\
%& \ \ \ \ \ \ \exp(- ([\svec',r, \gamma] - [\svec_i',r_i, \gamma_i])^\top \Hmat_{\svec',r,\gamma}^\inv ([\svec', r, \gamma] - [\svec'_i, r, \gamma_i]) )
%\end{align*}
%\normalsize
with covariance $\Hmat_{\svec',r,\gamma}$.
We set the covariance matrix $\Hmat_s = \ksize^\inv \Sigmamat_s$, where $\Sigmamat_s$ is a sample covariance,
and use a conditional covariance for $(\svec, r, \gamma)$.
% ADAM, TODO: I am leaving this intentionally vague, since we legitimately do not have space
% as described in Section \ref{sec_sampling}.
%More advanced approaches have been developed for conditional density estimation \citep{holmes2007fast}.
%%In this work, we used
%%a sample covariance matrix scaled by the number of points in the model, $\Hmat_s = \ksize^\inv \Sigma_s$
%%and a closed-form computation for conditional covariances.

%Using these prototypes, joint and conditional probabilities can be specified and sampled.
First consider a KDE model, for comparison, where all experience is used to define the distribution
\begin{equation*}
p_{\text{k}}(\svec, a, \svec', r, \gamma) = \tfrac{1}{\nsamples} {\textstyle \sum_{i=1}^\nsamples} k((\svec, a, \svec', r, \gamma), (\svec_i, a_i, \svec_i', r_i, \gamma_i))
\end{equation*}
This estimator puts higher density around more frequently observed transitions.
% if a transition is similar to stored transitions, then it will have a higher density value.
A conditional estimator is similarly intuitive, and also a consistent estimator \citep{holmes2007fast},
\small
\begin{align*}
N_{\text{k}}(\svec, a) &= \tfrac{1}{\nsamples} {\textstyle \sum_{i=1}^\nsamples} k_{\svec}(\svec, \svec_i) k_{a} (a, a_i)\\
\!p_{\text{k}}(\svec', r, \gamma | \svec, a) \!\!&=\!\!  \tfrac{1}{N_{\text{k}}(\svec, a)} \!\!\sum_{i=1}^\nsamples \!\! k_{\svec}(\svec, \!\svec_i) k_{a} (a, \!a_i) k_{\svec'\!,r\!,\gamma} (\!(\svec'\!,\! r\!,\!\gamma), \!(\svec_i',\!r_i,\!\gamma_i)\!)
\end{align*}
\normalsize
The experience $(\svec_i, a_i)$ similar to $(\svec, a)$ has higher weight in the conditional estimator: distributions centered at
$(\svec'_i, r_i, \gamma_i)$ contribute more to specifying $p(\svec', r, \gamma | \svec, a)$.
Similarly, it is straightforward to specify the conditional density $p(\svec | \svec', a)$.

When only prototype transitions are stored, joint and conditional densities can be similarly specified,
but prototypes must be weighted to reflect the density in that area.
We therefore need a method to select prototypes and to compute weightings.
Selecting representative prototypes or centers is a very active area of research, and we simply use a recent incremental and efficient algorithm
designed to select prototypes \cite{schlegel2017adapting}.
For the reweighting, however, we can design a more effective weighting exploiting the fact that we will only query the model using conditional distributions.

%We develop a new reweighting algorithm below, giving rise to our REM model. After that we describe how to sample from REMs and conclude this section with a high-level description of REM-Dyna in Algorithm \ref{alg_erwpspr}.

\myparagraph{Reweighting approach.}
%
%%%%%%%%%%%%%%% YANGCHEN STARTS %%%%%%%%%%%
%As our algorithm need to learn a set of prototypes, which is learned by using log determiniant criterion, we need to find a set of corresponding weights for us to correctly sample from the model. For example, given $(s, a)$, we need to sample $s', r, \gamma$ according to $p(s', r, \gamma | s, a) = \frac{p(s, a, s', r, \gamma)}{p(s, a)}$. And one challenge is, we are learning in an online manner and the prototypes are changing, hence we need a relatively stable and robust way of learning the weight. In the observance of only conditional weights are needed when sampling in Dyna methods, we propose the following stochastic optimization method to compute the weights. We use a generic notations to illustrate this algorithm first. Let $\Cmat_t \in \RR^{b \times d}$ be the prototypes at time step $t$, hence for $i$th line in the matrix $\zvec_i = (\xvec_i, \yvec_i)$, and we want to compute conditional weight vector $\cvec$ where each component is $c_i = p(\xvec_i | \yvec_i), \forall i \in \{1, 2, ..., b\}$. First, we propose the optimization objective as following.
%
%\begin{align*}
%\frac{1}{2} \min \sum_{i = 1}^{b} \sum_{t=1}^\nsamples  (c_i - k(\xvec_t,  \xvec_i))^2 k(\yvec_t, \yvec_i)
%\end{align*}
%
%Let $\yvec_i = (s_i, a_i), \xvec_i = (s'_i, r_i, \gamma_i)$ we can compute the corresponding conditional weight $c_i$.
%%%%%%%%%%%%%%% YANGCHEN ENDS %%%%%%%%%%%
%
We develop a reweighting scheme that takes advantage of the fact that Dyna only requires conditional models.
Because $p(\svec', r, \gamma | s, a) = p(\svec, a, \svec', r, \gamma)/p(\svec,a)$,
a simple KDE strategy is to estimate coefficients $p_i$ on the entire transition $(\svec_i, a_i, \svec'_i, r_i, \gamma_i)$
and $q_i$ on $(\svec_i, a_i)$, to obtain accurate densities $p(\svec, a, \svec', r, \gamma)$ and $p(\svec, a)$.
However, there are several disadvantages to this approach. The $p_i$ and $q_i$ need to constantly adjust, because the policy is changing. Further, when adding and removing prototypes incrementally, the other $p_i$ and $q_i$ need to be adjusted. Finally, $p_i$ and $q_i$ can get very small, depending on visitation frequency to a part of the environment, even if $p_i/q_i$ is not small.
Rather, by directly estimating the conditional coefficients $c_i = p(\svec'_i, r_i, \gamma_i | \svec_i, a_i)$, we avoid these problems. The distribution $p(\svec', r, \gamma | s, a)$ is stationary even with a changing policy; each $c_i$ can converge even during policy improvement and can be estimated independently of the other $c_i$.
%%The conditional density could then be computed as $p(\svec', r | s, a) = p(\svec, a, \svec', r)/p(\svec,a)$.
%%requiring only the coefficient $p_i/q_i$ to obtain the conditional weighting.
%However, computing $p_i$ and $q_i$ is problematic, because the policy is constantly changing and because they can get very small, depending on visitation frequency.
%%
%Rather, we can take advantage of the fact that the distribution $p(\svec', r | s, a)$ is in fact stationary, even with a changing policy,
%and directly estimate the conditional coefficients $c_i = p(\svec'_i, r_i, \gamma_i | \svec_i, a_i)$.
%This is further useful for a changing set of prototypes, since $p_i$ and $q_i$ would need to be adjusted when adding or removing prototypes,
%but $c_i$ does not.

 We can directly estimate $c_i$, because of the conditional independence assumption made by product kernels.
 To see why, for prototype $i$ in the product kernel in Equation \eqref{eq_product},
 %
 %\small
 \begin{align*}
 \!p(\svec, a, \svec', r, \gamma | \svec_i, a_i, \svec_i', r_i, \gamma_i)
 &=
 p(\svec, \!a | \svec_i, \!a_i)  p(\svec', \!r, \!\gamma | \svec_i', \!r_i, \!\gamma_i)
 \end{align*}
 %
% \normalsize
 Rewriting $p(\svec_i, \!a_i, \!\svec_i', \!r_i, \!\gamma_i) = c_i p(\svec_i, \!a_i)$
 %with {\small $c_i = p(\svec_i', r_i, \gamma_i | \svec_i, a_i)$}
 and because  $p(\svec, a | \svec_i, a_i) p(\svec_i, a_i) =  p(\svec_i, a_i | \svec, a) p(\svec, a)$,
% \begin{align*}
% p(\svec_i, a_i, \svec_i', r_i, \gamma_i) &= c_i p(\svec_i, a_i)\\
% p(\svec, a | \svec_i, a_i) p(\svec_i, a_i) &=  p(\svec_i, a_i | \svec, a) p(\svec, a)
% \end{align*}
% %and $p(\svec_i, a_i) = p(\svec_i, a_i | \svec, a) p(\svec, a)$.
%Consequently,
%for a sample $x = (\svec', r, \gamma, \svec, a)$ and prototype $x_i = (\svec_i, a_i, \svec_i', r_i, \gamma_i)$,
we can rewrite the probability as
 %
% \small
 \begin{align*}
p(\svec, \!a, \!\svec', \!r, \!\gamma) &= {\textstyle \sum_{i=1}^\ksize} [c_i p(\svec_i, \!a_i)]  p(\svec,\!a | \svec_i, \!a_i)  p(\svec', \!r, \!\gamma | \svec_i', r_i, \gamma_i)\\
& = {\textstyle \sum_{i=1}^\ksize} c_i p(\svec', r, \gamma | \svec_i', r_i, \gamma_i) p(\svec_i, a_i | \svec, a) p(\svec, a)
 \end{align*}
 %\normalsize
% giving
 %
 \begin{align*}
&\text{giving } \ \ \ \ \ \  p(\svec', r, \gamma | \svec, a) \\
  &= \tfrac{1}{p(\svec, a)} {\textstyle \sum_{i=1}^\ksize} c_i p(\svec', r, \gamma | \svec_i', r_i, \gamma_i)  p(\svec_i, a_i | \svec, a) p(\svec, a)\\
%&=  \sum_{i=1}^\ksize c_i p((\svec', r, \gamma) | (\svec_i', r_i, \gamma_i))  p(\svec_i, a_i | \svec, a) \\
&=  {\textstyle \sum_{i=1}^\ksize} c_i k_{\svec', r, \gamma}((\svec', r, \gamma), (\svec_i', r_i, \gamma_i))  k_{\svec} (\svec_i, \svec) k_a (a_i, a)
.
 \end{align*}
 \normalsize
%  % TOO LONG, editing above
% \begin{align*}
% &p((\svec, a, \svec', r, \gamma)) \\
% &= \sum_{i=1}^\ksize p((\svec, a, \svec', r, \gamma) | (\svec_i, a_i, \svec_i', r_i, \gamma_i)) p((\svec_i, a_i, \svec_i', r_i, \gamma_i))\\
% &= \sum_{i=1}^\ksize p((\svec', r, \gamma) | (\svec_i', r_i, \gamma_i)) c_i p(\svec_i, a_i | \svec, a) p(\svec, a)\\
%&  \implies p(\svec', r, \gamma | \svec, a) \\
%  &= \tfrac{1}{p(\svec, a)} \sum_{i=1}^\ksize c_i p((\svec', r, \gamma) | (\svec_i', r_i, \gamma_i))  p(\svec_i, a_i | \svec, a) p(\svec, a)\\
%&=  \sum_{i=1}^\ksize c_i p((\svec', r, \gamma) | (\svec_i', r_i, \gamma_i))  p(\svec_i, a_i | \svec, a) \\
%&=  \sum_{i=1}^\ksize c_i k_{\svec', r, \gamma}((\svec', r, \gamma), (\svec_i', r_i, \gamma_i))  k_{\svec} (\svec_i, \svec) k_a (a_i, a)
% \end{align*}
% %

Now we simply need to estimate $c_i$.
Again using the conditional independence property, we can prove the following.
%\vspace{-1cm}
\begin{minipage}{0.483\textwidth}
\begin{theorem}\label{thm_c}
Let $\rho(t,i) \defeq k_{\svec,a}((\svec_t, a_t),(\svec_i, a_i))$ be the similarity of $\svec_t, a_t$ for sample $t$ to $\svec_i, a_i$ for prototype $i$. Then
\begin{align*}
c_i \!=\! \argmin_{c_i \ge 0} {\textstyle \sum_{t=1}^\nsamples}  (c_i - k_{\svec', r, \gamma}(\!(\svec'_t, r_t, \gamma_t),(\svec'_i, r_i, \gamma_i)\!))^2 \rho(t,i)
\end{align*}
is a consistent estimator of $p(\svec_i', r_i, \gamma_i | \svec_i, a_i)$.
 \end{theorem}
 The proof for this theorem, and a figure demonstrating the difference between KDE and REM, are provided in the appendix.
 \end{minipage}
%The conditional weight $c_i$ reflects how probable it is to observe $\svec_i', r_i, \gamma_i$, given $\svec_i, a_i$,
%and so can be updated based on observed transitions.
%We can estimate such conditional weightings by minimizing
%%
%\begin{align*}
%\min_{c_i \ge 0} \sum_{t=1}^\nsamples  (c_i - k_{\svec', r, \gamma}((\svec'_t, r_t, \gamma_t),(\svec'_i, r_i, \gamma_i)))^2 \rho(t,i)
%\end{align*}
%%
%where $\rho(t,i)$ reflects the similarity of $\svec_t, a_t$ for sample $t$ to $\svec_i, a_i$ for prototype $i$: $\rho(t,i) = k_{\svec,a}((\svec_t, a_t),(\svec_i, a_i))$.
%%
%%\begin{align*}
%%c_i = \frac{ \sum_{t=1}^\nsamples  \rho(t,i) k((\svec'_t, r_t, \gamma_t),(\svec'_i, r_i, \gamma_i)) }{ \sum_{t=1}^\nsamples  \rho(t,i)}
%%\end{align*}
%%
% MARTHA: Fricken latex. I am just putting minipages everywhere to get the spacing to be reasonable
\begin{minipage}{0.483\textwidth}
\vspace{-0.2cm}
Though there is a closed form solution to this objective, we use an incremental stochastic update to avoid storing additional variables and for the model to be more adaptive. For each transition, the $c_i$ are updated for each prototype as
\end{minipage}
\begin{equation*}
c_i  \gets (1- \rho(t,i)) c_i + \rho(t,i) k_{\svec', r, \gamma}((\svec'_t, r_t, \gamma_t),(\svec'_i, r_i, \gamma_i))
\end{equation*}
%
%For a new prototype, its weighting is initialized as $c_i = 1$.
%and let further interaction decrease the probability. This initialization is unlikely to be damaging for a non-stationary world, because
%that transition was just observed and so is likely to be valid.

The resulting REM model is
%
%The resulting KDE conditional model is
%\small
\begin{align*}
\beta_i(\svec, a) &\defeq \tfrac{c_i}{N(\svec, a)} \ k_{\svec}(\svec, \svec_i) k_{a}(a, a_i) \\
&\text{ where } N(\svec, a) \defeq {\textstyle \sum_{i=1}^\ksize} c_i \ k_{\svec}(\svec, \svec_i) k_{a}(a, a_i)\\
p(\svec', r, \gamma | \svec, a) &\defeq {\textstyle \sum_{i=1}^\ksize} \beta_i(\svec, a)  k_{\svec',r,\gamma}((\svec', r, \gamma), (\svec_i', r_i, \gamma_i))
.
\end{align*}
%
%\normalsize
To sample predecessor states, with $p(\svec | \svec', a)$, the same set of $\ksize$ prototypes can be used,
with a separate set of conditional weightings estimated as
$c^{\text{r}}_i \gets (1- \rho^{\text{r}}(t,i)) c^{\text{r}}_i + \rho^{\text{r}}(t,i) k_{\svec}(\svec,\svec_i)$
for $\rho^{\text{r}}(t,i) \defeq k_{\svec}(\svec',\svec'_i) k_{a}(a, a_i)$.

%\subsection{Sampling from REMs}\label{sec_sampling}
\myparagraph{Sampling from REMs.}
Conveniently, to sample from the REM conditional distribution, the similarity across next states and rewards need not be computed.
Rather, only the coefficients $\beta_i(\svec, a)$ need to be computed. A prototype is sampled with probability $\beta_i(\svec, a)$; if prototype $j$ is sampled, then the density (Gaussian) centered around $(\svec_j', r_j, \gamma_j)$ is sampled.

In the implementation, the terms $(2\pi)^{-\xdim/2} | \Hmat_s |^{-1/2}$ in the Gaussian kernels are omitted, because as fixed constants they can be normalized out. All kernel values then are in $[0,1]$, providing improved numerical stability and the straightforward initialization $c_i=1$ for new prototypes. REMs are linear in the number of prototypes, for learning and sampling, with complexity per-step independent of the number of samples.
%, which can then only have a maximum value of 1.

\myparagraph{Addressing issues with scaling with input dimension.}
In general, any nonnegative kernel $k_\svec(\cdot, \svec)$ that integrates to one is possible. There are realistic low-dimensional physical systems for which Gaussian kernels have been shown to be highly effective, such as in robotics \citep{deisenroth2011pilco}. Kernel-based approaches can, however, extend to high-dimensional problems with specialized kernels. For example, convolutional kernels for images have been shown to be competitive with neural networks \citep{mairal2014convolutional}. Further, learned similarity metrics or embeddings enable data-driven models---such as neural networks---to improve performance, by replacing the Euclidean distance. This combination of probabilistic structure from REMs and data-driven similarities for neural networks is a promising next step.

\begin{figure*}[t]
	% \vspace{-1.2cm}
	\centering
	\includegraphics[width=0.80\textwidth]{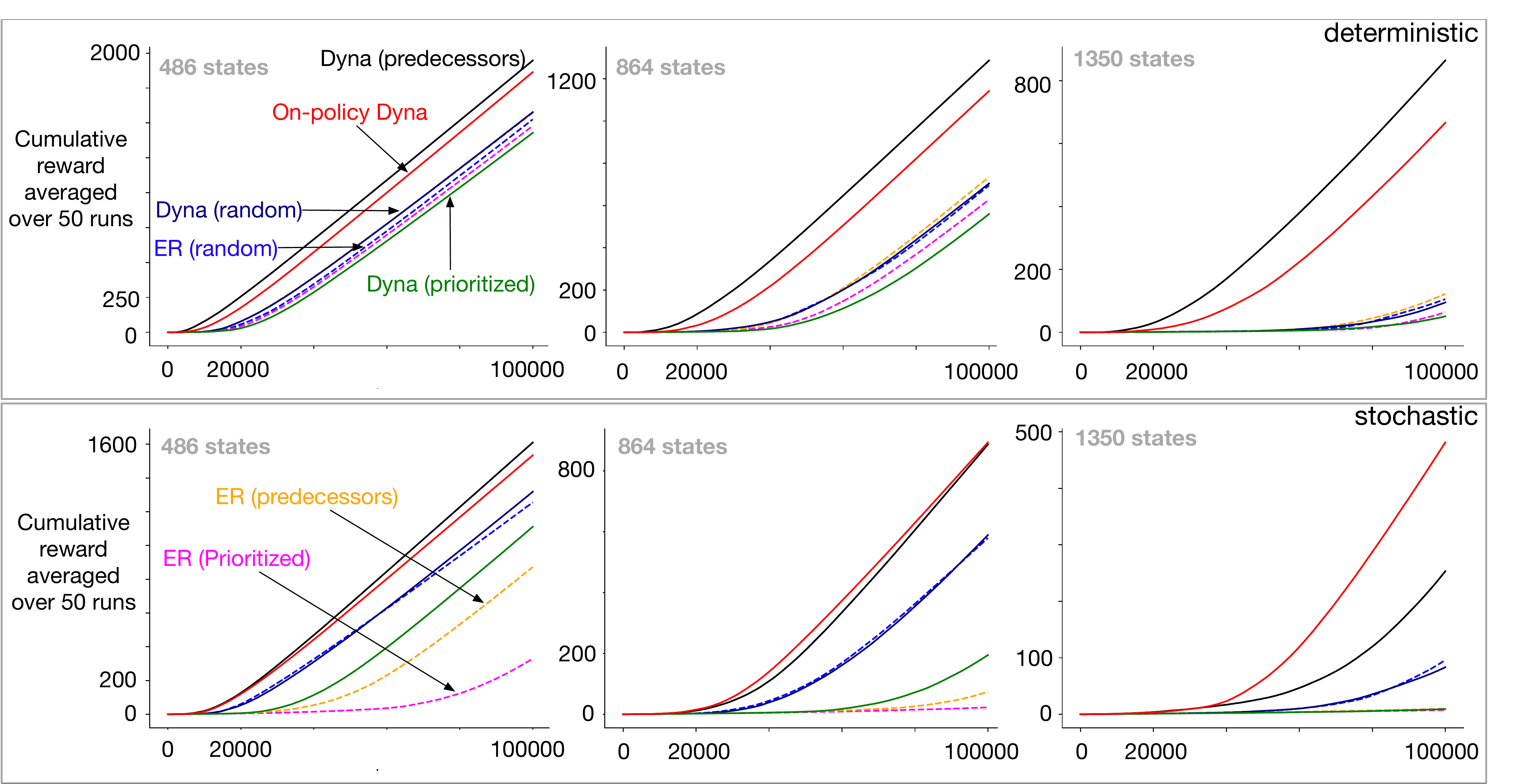} \label{fig:grid486 states}
%	\subfigure[Deterministic]{
%		\includegraphics[width=0.9\textwidth]{figures/tabularDet} \label{fig:grid486 states}} \\
%		\vspace{-0.25cm}
%	\subfigure[Stochastic]{
%		\includegraphics[width=0.9\textwidth]{figures/tabularStoc} \label{fig:grid486 states}}
%	\subfigure[864 states, Deterministic]{
%		\includegraphics[width=\figwidththree]{figures/tabular/states_864/deterministic.png} \label{fig:grid864states}}
%	\subfigure[1350 states, Deterministic]{
%		\includegraphics[width=\figwidththree]{figures/tabular/states_1350/deterministic.png}\label{fig:grid1350states}}
%	\subfigure[486 states, Stochastic]{
%		\includegraphics[width=\figwidththree]{figures/tabular/states_486/stochastic.png} \label{fig:grid486states_st}}
%	\subfigure[864 states, Stochastic]{
%		\includegraphics[width=\figwidththree]{figures/tabular/states_864/stochastic.png} \label{fig:grid864states_st}}
\	%\begin{minipage}{\figwidththree}
	\vspace{-0.35cm}
	\caption{
		Learning curves for a varying number of states, for the deterministic (upper) and stochastic (lower) gridworlds. The x-axis shows the number of interactions with the environment}\label{fig:gridworldexp}
	%\end{minipage}
	\vspace{-0.25cm}
\end{figure*}

\section{Experiments}

%In this section, we empirically investigate the design choices for ER's buffer and Dyna's search-control queue
%as well as investigate the utility of the proposed model-learning technique, for REM-Dyna in the function approximation setting.
%We highlight that Dyna can significantly improve on ER, with the effect magnified for stochastic and larger domains,
%and that the conclusions persist even when learning a REM model for Dyna for continuous states.

We first empirically investigate the design choices for ER's buffer and Dyna's search-control queue in the tabular setting. Subsequently, we examine the utility of REM-Dyna, our proposed model-learning technique, by comparing it with ER and other model learning techniques in the function approximation setting. Maintaining the buffer or queue involves determining how to add and remove samples, and how to prioritize samples. All methods delete the oldest samples. Our experiments (not shown here), showed that deleting samples of lowest priority---computed from TD error---is not effective in the problems we studied. We investigate three different settings:\\
%We investigate four different approaches for maintaining the buffer and search-control queues, with entire transitions stored for ER and only $(\svec, a)$ stored for Dyna:\\
1) \textit{Random}: samples are drawn randomly.\\
2) \textit{Prioritized}: samples are drawn probabilistically according to the absolute TD error of the transitions \cite[Equation 1]{schaul2016prioritized} (exponent  = 1).\\
3) \textit{Predecessors}: same as \textit{Prioritized}, and predecessors of the current state are also added to the buffer or queue.
%Samples from the buffer or queue are chosen probabilistically according to their priority, using stochastic prioritization proposed in \cite[Equation 1]{schaul2016prioritized} (exponent  = 1).
%TheER ntire transitions stored for ER and only $(\svec, a)$ stored for Dyna
%We first empirically investigate the design choices for ER’s buffer and Dyna’s search-control queue in the tabular setting. Subsequently, we examine the utility of REM-Dyna, our proposed model-learning technique, by comparing it with ER and other model learning techniques in the function approximation setting. Maintaining the buffer or queue involves determining how to add and remove samples, and how to prioritize samples for replay.  We investigate four different approaches for maintaining the buffer and search-control queues, with entire transitions stored for ER and only $(\svec, a)$ stored for Dyna:\\
%1) \textit{Recency}: All samples have a uniform priority; the oldest sample is removed.\\
%2) \textit{Prioritized}: Samples have a priority based on absolute TD-error; the lowest priority sample is removed.\\
%3) \textit{Prioritized-Recency}: Samples have a priority based on absolute TD-error; the oldest sample is removed.\\
%4) \textit{Priorities \& Predecessors}: An extension of \textit{Prioritized-Recency} in which after every replay step, the predecessors of the current state are also added to the buffer or queue.\\
%Samples from the buffer or queue are chosen probabilistically according to their priority, using stochastic prioritization proposed in \cite[Equation 1]{schaul2016prioritized} (exponent  = 1).

We also test using \emph{On-policy} transitions for Dyna, where only $\svec$ is stored on the queue and actions simulated according to the current policy;
%and when $\svec$ is drawn from the queue, the action is simulated according to the current policy;
the queue is maintained using priorities and predecessors.
In Dyna, we use the learned model to sample predecessors $\svec$ of the current $\svec'$, for all actions $a$, and add them to the queue.  In ER, with no environment model, we use a simple heuristic which adds the priority of the current sample to the preceding sample in the buffer \citep{schaul2016prioritized}.
Note that \citet{vanseijen2015adeeper} relate Dyna and ER, but specifically for a theoretical equivalence in policy evaluation based on a non-standard form of replay related to true online methods, and thus we do not include it.

\myparagraph{Experimental settings}:
All experiments are averaged over many independent runs, with the randomness controlled based on the run number.
All learning algorithms use $\epsilon$-greedy action selection ($\epsilon = 0.1$) and Q-learning to update the value function in both \textit{learning} and \textit{planning} phases. The step-sizes are swept in $[0.1, 1.0]$. The size of the search-control queue and buffer was fixed to 1024---large enough for the micro-worlds considered---and the number of planning steps was fixed to 5.

A natural question is if the conclusions from experiments in the below microworlds extend to larger environments. Microworlds are specifically designed to highlight phenomena in larger domains, such as creating difficult-to-reach, high-reward states in River Swim described below.
The computation and model size are correspondingly scaled down, to reflect realistic
limitations when moving to larger environments. The trends obtained when varying the size and stochasticity of these environments provides insights into making such changes in larger environments. Experiments, then, in microworlds enable a
\begin{minipage}{0.483\textwidth}
\vspace{-0.55cm}
more systematic issue-oriented investigation and suggest directions for further investigation for use in real domains.

\myparagraph{Results in the Tabular Setting}:
% deterministic and stochastic Tabular results in Maze Navigation}
%Dyna and ER have largely been empirically explored for the (deterministic) tabular setting, with the exception of Prioritized ER \citep{schaul2016prioritized}.
To gain insight into the differences between Dyna and ER, we first consider them in the deterministic and stochastic variants of a simple gridworld with increasing state space size.  ER has largely been explored in deterministic problems,
%(including Prioritized ER \citep{schaul2016prioritized}),
and most work on Dyna has only considered the
tabular setting.
The gridworld is discounted with $\gamma = 0.95$, and episodic with obstacles and one goal, with a reward of 0 everywhere except the transition into goal, in which case the reward is +100. The agent can take four actions.
% which deterministically move the agent to an adjacent state, where
In the stochastic variant each action takes the agent to the intended next state with probability 0.925, or one of the other three adjacent states with probability 0.025. In the deterministic setting, Dyna uses a table to store next state and reward for each state and action; in stochastic, it estimates the probabilities of each observed transition via transition counts.
\end{minipage}

 \begin{figure*}[t]
	% \vspace{-1.2cm}
	\centering
	\includegraphics[width=1.0\textwidth]{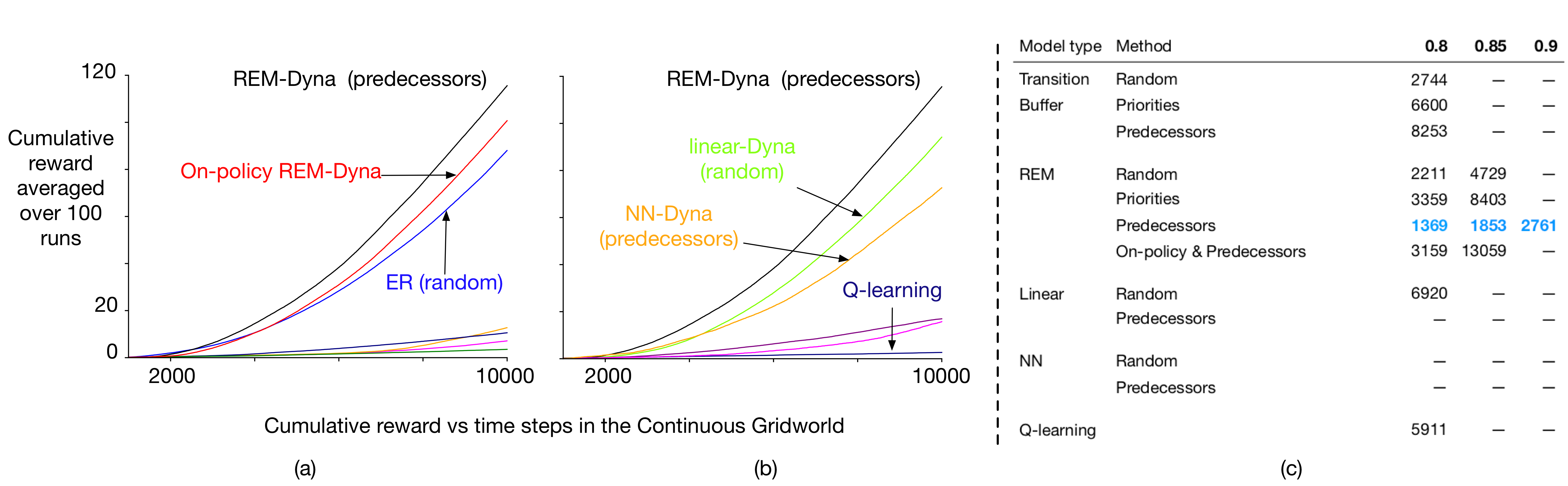}
%	\subfigure[Deterministic]{
%		\includegraphics[width=0.9\textwidth]{figures/tabularDet} \label{fig:grid486 states}} \\
%		\vspace{-0.25cm}
%	\subfigure[Stochastic]{
%		\includegraphics[width=0.9\textwidth]{figures/tabularStoc} \label{fig:grid486 states}}
%	\subfigure[864 states, Deterministic]{
%		\includegraphics[width=\figwidththree]{figures/tabular/states_864/deterministic.png} \label{fig:grid864states}}
%	\subfigure[1350 states, Deterministic]{
%		\includegraphics[width=\figwidththree]{figures/tabular/states_1350/deterministic.png}\label{fig:grid1350states}}
%	\subfigure[486 states, Stochastic]{
%		\includegraphics[width=\figwidththree]{figures/tabular/states_486/stochastic.png} \label{fig:grid486states_st}}
%	\subfigure[864 states, Stochastic]{
%		\includegraphics[width=\figwidththree]{figures/tabular/states_864/stochastic.png} \label{fig:grid864states_st}}
\	%\begin{minipage}{\figwidththree}
		\vspace{-0.7cm}
	\caption{
		(a) compares variants of ER and REM-Dyna. REM-Dyna with predecessor states and Random ER accumulate significantly more reward than all other variants, with REM-Dyna statistically significantly better (non-overlapping confidence intervals) than ER by the end of the run. (b) shows Dyna with different models.
		%with different methods for managing the search-control queue.
REM-Dyna is statistically significantly better that NN-Dyna and Linear-Dyna. For NNs and REMs, using predecessors
		%REM-Dyna, linear-Dyna with predecessor states, and Dyna with a NN model and predecessor states
		is significantly better, unlike Linear-Dyna which learns inaccurate models. (c) Results on River Swim, with number of steps required to obtain a ratio of 80\%, 85\% and 90\% between the cumulative reward for the agent relative to the cumulative reward of the optimal policy. If there is no entry, then the agent was unable to achieve that performance within the 20,000 learning steps.}\label{fig:faexp_conti_gridworld}
	%\end{minipage}
	\vspace{-0.2cm}
\end{figure*}

Figure \ref{fig:gridworldexp} shows the reward accumulated by each agent over $100,000$ time-steps. We observe that: 1) Dyna with priorities and predecessors outperformed all variants of ER, and the performance gap increases with gridworld size. 2) TD-error based prioritization on Dyna's search control queue improved performance only when combined with the addition of predecessors;
otherwise, unprioritized variants outperformed prioritized variants. We hypothesize that this could be due to out-dated priorities, previously suggested to be problematic \citet{peng1993efficient,schaul2016prioritized}. 3) ER with prioritization performs slightly worse than unprioritized ER variants for the deterministic setting, but its performance degrades considerably in the stochastic setting. 4) On-Policy Dyna with priorities and predecessors outperformed the regular variant in the stochastic domain with a larger state space. 5) Dyna with similar search-control strategies to ER, such as recency and priorities, does not outperform ER; only with the addition of improved search-control strategies is there an advantage. 6) Deleting samples from the queue or transitions from the buffer according to recency was always better than deleting according to priority for both Dyna and ER.

\myparagraph{Results for Continuous States.}
We recreate the above experiments for continuous states, and additionally explore the utility of REMs for Dyna. We compare to using a Neural Network model---with two layers, trained with the Adam optimizer on a sliding buffer of 1000 transitions---and to a Linear model predicting features-to-expected next features rather than states, as in Linear Dyna. We improved upon the original Linear Dyna by learning a reverse model and sweeping different step-sizes for the models and updates to $\hat{q}_{\pi}$.
%We use to Linear Dyna as a baseline, which is the only other Dyna-style algorithm compatible with function approximation. As motivated in Section \ref{sec_generative}, other model choices could be incorporated into Dyna, but currently do not satisfy the outlined criteria. Rather, here we focus on whether
%our conclusions from the tabular experiments extend to continuous state using our new lightweight, data-efficient model, REM.

We conduct experiments in two tasks: a Continuous Gridworld and River Swim. Continuous Gridworld is a continuous variant of a domain introduced by \citep{peng1993efficient}, with $(x, y) \in [0, 1]^2$, a sparse reward of 1 at the goal, and a long wall with a small opening.
%Actions move the agent stochastically up, down, left, right by 0.05 units, with a variance of 0.01, and with probability 0.1 the action fails, where the environment executes a random action.
Agents can choose to move 0.05 units up, down, left, right, which is executed successfully with probability $0.9$ and otherwise the environment executes a random move. Each move has noise $\mathcal{N}(0, 0.01)$. River Swim is a difficult exploration domain, introduced as a tabular domain \cite{strehl2008modeliemdp}, as a simple simulation of a fish swimming up a river. We modify it to have a continuous state space $[0, 1]$. On each step, the agent can go right or left, with the river pushing the agent towards the left. The right action succeeds with low probability depending on the position, and the left action always succeeds. There is a small reward $0.005$ at the leftmost state (close to $0$), and a relatively large reward $1.0$ at the rightmost state (close to $1$). The optimal policy is to constantly select right.
Because exploration is difficult in this domain, instead of $\epsilon$-greedy, we induced a bit of extra exploration by initializing the weights to $1.0$. For both domains, we use a coarse tile-coding, similar to state-aggregation.

%We conduct experiments in two tasks: Block World and River Swim. The results are showed in Figure~\ref{fig:faexp}. Block World is a continuous variant of the gridworld problem, with sparse reward, but with a long wall with a hole, with observations $(x, y) \in [0, 1]^2$. Action moves the agent stochastically up, down, left, right by 0.05 units, with a variance of 0.01. River Swim is a difficult exploration domain, as a simple simulation of a fish swimming up a river. On each step, the agent can can right or left, with the river pushing the agent towards the left. The right action succeeds with probability 0.1, and the left action succeeds with probability 1. There is a small positive reward at the bottom of the river---the leftmost state---and a large positive value at the top---the rightmost state. The optimal policy is to constantly select right.
%Because exploration is difficult in this domain, in addition to $\epsilon$-greedy, we induced a bit of extra exploration by initializing the weights to $1.0$.
%
%From results of both domains, we see that our REM-Dyna algorithm achieves best performance, which shows our model and the conditional weighing scheme are robust. Importantly, it further verifies the conclusion from the tabular case where the predecessor state indeed plays an important role in achieving nice performance. For all buffer based algorithms, recency buffer plays a more effective role, especially in stochastic environment.

REM-Dyna obtains the best performance on both domains, in comparison to the ER variants and other model-based approaches.
For search-control in the continuous state domains, the results in Figures \ref{fig:faexp_conti_gridworld} parallels the conclusions from the tabular case.
For the alternative models, REMs outperform both Linear models and NN models. For Linear models, the model-accuracy was quite low and the step-size selection sensitive. We hypothesize that this additional tuning inadvertently improved the Q-learning update, rather than gaining from Dyna-style planning;
in River Swim, Linear Dyna did poorly.
%with less carefully chosen step-sizes, Linear Dyna performed significantly worse.
Dyna with NNs performs poorly because the NN model is not data-efficient; after several 1000s of more learning steps, however, the model does finally become accurate. This highlights the necessity for data-efficient models, for Dyna to be effective. In Riverswim, no variant of ER was within 85\% of optimal, in 20,000 steps, whereas all variants of REM-Dyna were, once again particularly for REM-Dyna with Predecessors.

\section{Conclusion}

In this work, we developed a semi-parametric model learning approach, called Reweighted Experience Models (REMs),
for use with Dyna for control in continuous state settings. We revisited a few key dimensions for maintaining the search-control queue for Dyna,
to decide how to select states and actions from which to sample. These included understanding the importance of using recent samples,
prioritizing samples (with absolute TD-error), generating predecessor states that lead into high-priority states, and generating on-policy transitions.
%We additionally considered the utility of simulating on-policy transitions, storing only states in the search-control queue, rather than storing the state and action and simulating from that previously taken action.
We compared Dyna to the simpler alternative, Experience Replay (ER), and considered similar design decisions for its transition buffer. We highlighted several criteria for the model to be useful in Dyna, for one-step sampled transitions, namely being data-efficient, robust to forgetting, enabling conditional models and being efficient to sample. We developed a new semi-parametric model, REM, that uses similarities to a representative set of prototypes, and requires only a small set of coefficients to be learned. We provided a simple learning rule for these coefficients, taking advantage of a conditional independence assumption and that we only require conditional models.
%We show how these coefficients can be more simply learned, by taking advantage of a conditional independence assumption in our model and the fact that we only require conditional models.
We thoroughly investigate the differences between Dyna and ER, in several microworlds for both tabular and continuous states,
showing that Dyna can provide significant gains through the use of predecessors and on-policy transitions. We further highlight that REMs are an effective model for Dyna, compared to using a Linear model or a Neural Network model.

{
%\scriptsize
\newpage
\small
\bibliography{paper.bib}
\bibliographystyle{named}
}

%!TEX root = paper.tex

\appendix

\newcommand{\Smat}{\mathbf{S}}
\section{Consistency of conditional probability estimators}

 \textbf{Theorem \ref{thm_c}}
Let $\rho(t,i) = k_{\svec,a}((\svec_t, a_t),(\svec_i, a_i))$ be the similarity of $\svec_t, a_t$ for sample $t$ to $\svec_i, a_i$ for prototype $i$. Then
\begin{align*}
c_i = \argmin_{c_i \ge 0} \sum_{t=1}^\nsamples  (c_i - k_{\svec', r, \gamma}((\svec'_t, r_t, \gamma_t),(\svec'_i, r_i, \gamma_i)))^2 \rho(t,i)
\end{align*}
is a consistent estimator of $p(\svec_i', r_i, \gamma_i | \svec_i, a_i)$. 
\begin{proof}
The closed-form solution for this objective is
\begin{align*}
c_i &= \frac{ \sum_{t=1}^\nsamples  \rho(t,i) k_{\svec', r,\gamma}((\svec'_t, r_t, \gamma_t),(\svec'_i, r_i, \gamma_i)) }{ \sum_{t=1}^\nsamples  \rho(t,i)} \\
&= \frac{ \tfrac{1}{\nsamples}\sum_{t=1}^\nsamples  \rho(t,i) k_{\svec', r,\gamma}((\svec'_t, r_t, \gamma_t),(\svec'_i, r_i, \gamma_i))}{ \tfrac{1}{\nsamples}\sum_{t=1}^\nsamples  \rho(t,i)}
\end{align*}
As $\nsamples \rightarrow \infty$, 
\begin{align*}
&\lim_{T\rightarrow\infty} \frac{ \tfrac{1}{\nsamples}\sum_{t=1}^\nsamples  \rho(t,i) k_{\svec', r,\gamma}((\svec'_t, r_t, \gamma_t),(\svec'_i, r_i, \gamma_i))}{ \tfrac{1}{\nsamples}\sum_{t=1}^\nsamples  \rho(t,i)} \\
&= \frac{\lim_{T\rightarrow\infty} \ \ \tfrac{1}{\nsamples} \sum_{t=1}^\nsamples  \rho(t,i) k_{\svec', r,\gamma}((\svec'_t, r_t, \gamma_t),(\svec'_i, r_i, \gamma_i))}{\lim_{T\rightarrow\infty} \ \ \tfrac{1}{\nsamples}\sum_{t=1}^\nsamples  \rho(t,i)} \\
&= \frac{\E[k_{\svec,a}((\Smat, A),(\svec_i, a_i))k_{\svec', r,\gamma}((\Smat', R, \gamma),(\svec'_i, r_i, \gamma_i))]}{\E[k_{\svec,a}((\Smat,A),(\svec_i, a_i))]}
\end{align*}
with expectation according to $p(\svec, a, \svec', r, \gamma)$. 
The second equality holds because (a) all three limits exist and (b) the limit of the denominator is not zero: $\E[k_{\svec,a}((\Smat,A),(\svec_i, a_i))] \neq 0$. 

Expanding out these expectations, where $k_{\svec,a}((\svec,a),(\svec_i, a_i)) = p(\svec, \avec | \svec_i, a_i) = p(\svec_i, a_i | \svec, a)$ by the symmetry of the kernel, we get 
\begin{align*}
\E[k_{\svec,a}((\Smat,A),(\svec_i, a_i))] 
&= \int_{\States} \sum_a p(\svec_i, \avec_i | \svec, a) p(\svec, a) d(s,a) \\
&= p(\svec_i, \avec_i )
\end{align*}
Similarly, 
\begin{align*}
&\E[k_{\svec,a}((\Smat, A),(\svec_i, a_i))k_{\svec', r,\gamma}((\Smat', R, \gamma),(\svec'_i, r_i, \gamma_i))]\\
&= \E[p(\svec_i, a_i | \Smat, A) p(\svec_i', r_i, \gamma_i | \Smat', R, \gamma) ]\\
&= \E[p(\svec_i, a_i, \svec_i', r_i, \gamma_i | \Smat, A, \Smat', R, \gamma) ]\\
&= p(\svec_i, a_i, \svec_i', r_i, \gamma_i)
\end{align*}
Therefore, 
\begin{align*}
&\frac{\E[k_{\svec,a}((\Smat, A),(\svec_i, a_i))k_{\svec', r,\gamma}((\Smat', R, \gamma),(\svec'_i, r_i, \gamma_i))]}{\E[k_{\svec,a}((\Smat,A),(\svec_i, a_i))]}\\
&= \frac{p(\svec_i, a_i, \svec_i', r_i, \gamma_i)}{p(\svec_i, \avec_i )}\\
&= p(\svec_i', r_i, \gamma_i | \svec_i, a_i)
\end{align*}
and so $c_i$ converges to $p(\svec_i', r_i, \gamma_i | \svec_i, a_i)$ as $\nsamples \rightarrow \infty$. 
\end{proof}

\section{REM Algorithmic Details}\label{app_rem}
\begin{figure*}[t]
	\centering
	\subfigure[Kernel Density Estimator (KDE)]{
			\includegraphics[width=0.45\textwidth]{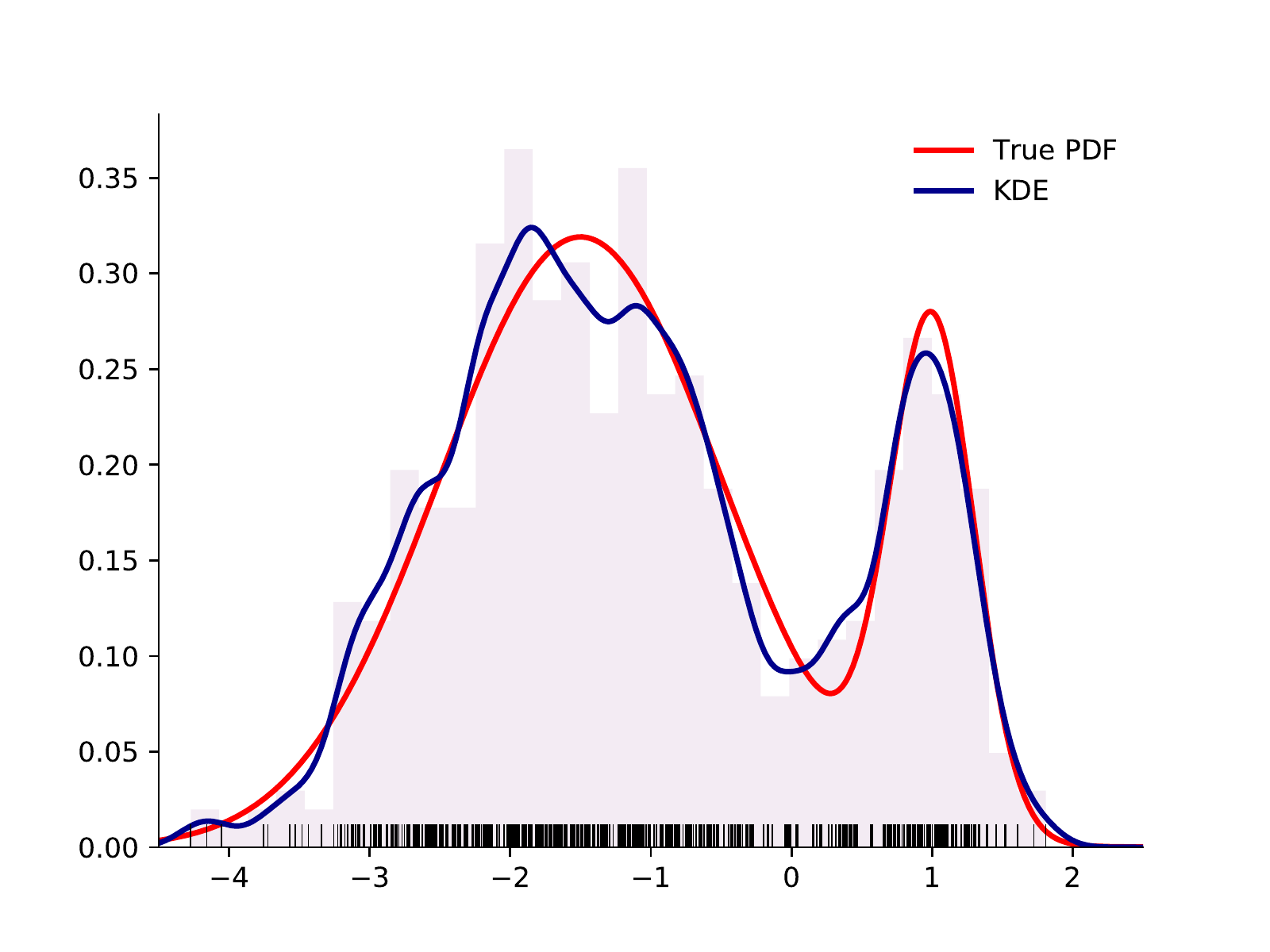} \label{fig_kde}} 
	\subfigure[Reweighted Experience Model (REM)]{
			\includegraphics[width=0.45\textwidth]{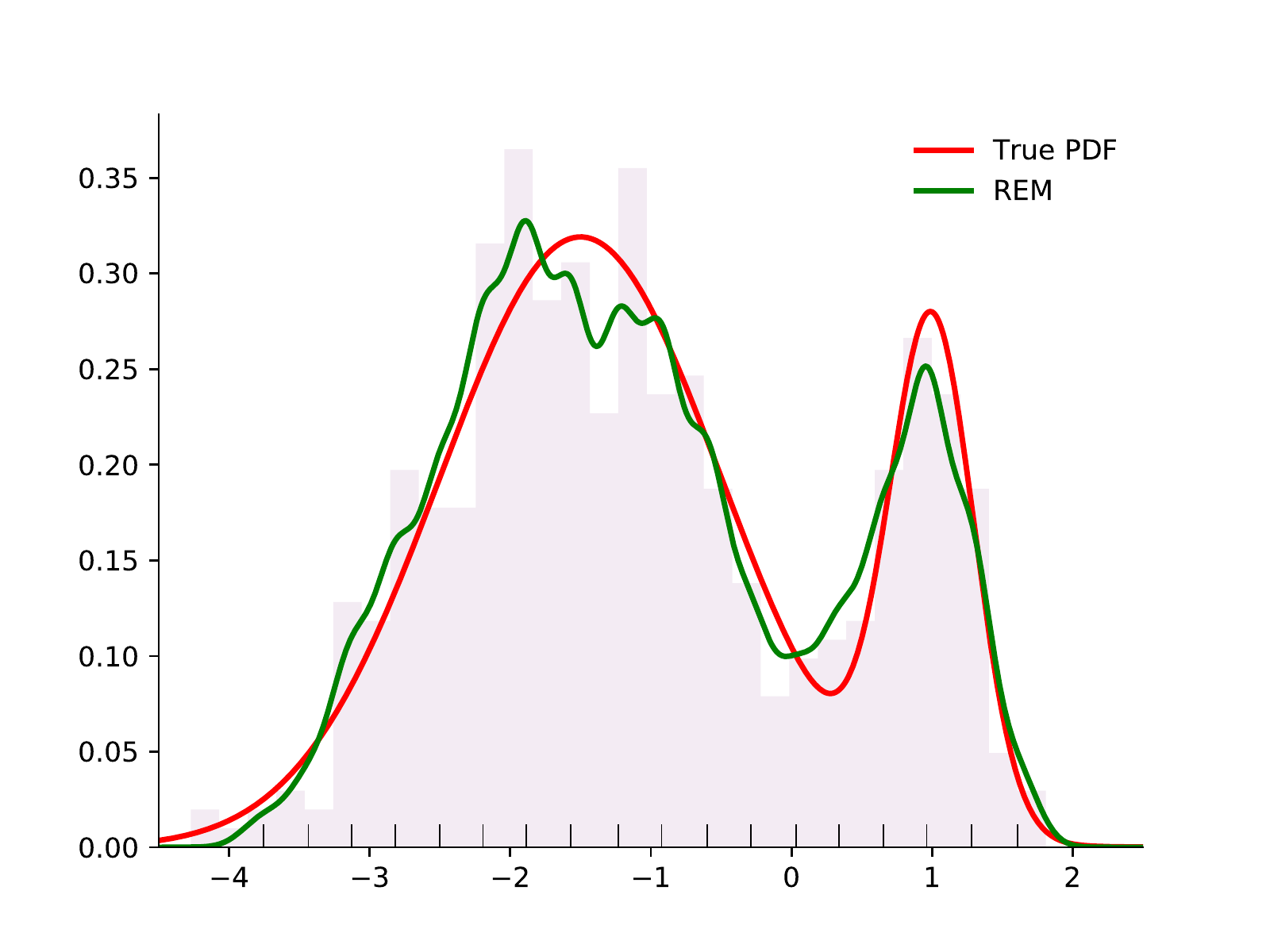} \label{fig_rem}} 			
	\caption{ This figure highlights the difference between (a) KDE and (b) REM. The KDE uses all data observed, represented by the ticks on the x-axis. The density is a mixture model with each data point as a center, and uniform weightings on points. REM, on the other hand, subselects a representive subset of the data as centers, and then uses the reweighting scheme described to adjust the coefficients. The distinction between REM and a standard Gaussian Mixture Model is in how it subselects points and how it computes the coefficients; otherwise, it can be seen as a Gaussian Mixture Model in this figure. For both KDE and REM, we set the variance to $0.15$ in the RBFs. The total number of samples is $500$, with REM sub-selecting $30$. The data is generated from a bi-modal distribution (black), with sample density depicted with the histogram (lavender).}\label{fig_kderem} 
	%\end{minipage}
\end{figure*}

Algorithm \ref{alg_erwpspr} summarizes REM-Dyna, our online algorithm for learning, acting, and sample-based planning.
Supporting pseudocode, for sampling and updating REMs, is given in Section \ref{app_pseudocode} below. We include Experience Replay with Priorities in Algorithm \ref{alg_er}, for comparison. 
We additionally include a diagram highlighting the difference between KDE and REM, to approximate densities, in Figure \ref{fig_kderem}.

For the queue and buffer, we maintain a circular array. When a sample is added to the array, with priority $P$, it is placed in the spot with the oldest transition. When a state-action or transition is sampled with priority $P$ from the array, it is used to update the weights and its priority in the array is updated with its new priority. Therefore, it is not removed from the array, simply updated. Array elements are only removed once they are the oldest, implemented by incrementing the index each time a new point is added to the array. 

\begin{algorithm}[t]
\caption{REM-Dyna with Predecessors and Prioritization}
\label{alg_erwpspr}
\begin{algorithmic}
%\State $\mathcal{M}$ is a REM model, $Q$ is the action value function, $q$ is priority queue, 
\State $n$ is the number of planning steps, $\alpha$ is Q-learning stepsize, $B$ is search-control queue and $\mathcal{M}$ is the model, $f$ is the branching factor, $\epsilon_p$ is the priority threshold
%$\beta\in\mathbb{R}^+$ the threshold for adding transitions to the priority queue, %and $\tau\in \mathbb{R}^+$ be a threshold on the similarity between a prototype in our model and a state in the priority queue, 
%$\hat{q}$ is the learned q-function % parameterized by $\theta\in\mathbb{R}^n$.
%\State $\triangleright$ Input
\While {true}
\State Get transition $(\svec, a, \svec', r, \gamma)$
\State Q-learning update with transition using stepsize $\stepsize$ % using Sarsa or Q-learning
%\State $S \leftarrow$ current non-terminal state
%\State $A \leftarrow$ policy(S,$\hat{q}$) 
%\State Execute $A$, observe next state $S'$ and reward $R$
\State Update REM $ \mathcal{M}$ with $(\svec,a,\svec',r, \gamma)$ (Algorithm \ref{alg_remupdate})
\State $P \gets | r + \gamma \max_{a'} \hat{q}(\svec',a') - \hat{q}(\svec,a)|$
\State Insert $(\svec,a)$ into $B$ with priority $P + \epsilon_p$
\For {$n$ times}
\State $(\svec,a) \gets$ sample from $B$ by priority
\State Sample $(\svec', r, \gamma) \sim \mathcal{M}(\svec,a)$ 
\State Q-learning update with stepsize $\frac{\stepsize}{\sqrt{n}}$ 
\State Update priority of $(\svec, a, \svec', r, \gamma)$
%\State $\triangleright$ obtain $\bar{\svec}, \bar{a}$ leading to $\svec$
\For {$f$ times} % (Algorithm \ref{alg_sampling_from_REM_reverse})
\State $\bar{\svec}, \bar{a} \sim$  {\small SamplePredecessors}($\svec, \mathcal{M}$)
\State sample $(\bar{\svec}', \bar{r}, \bar{\gamma}) \sim \mathcal{M}(\bar{\svec}, \bar{a})$ (Algorithm \ref{alg_sampling_from_REM_forward})
\State $P \gets |\bar{r} + \bar{\gamma} \max_{a'} \hat{q}(\bar{\svec}',a') - \hat{q}(\bar{\svec},\bar{a})|$ 
%\If{$P \ge \epsilon_p$}
\State Insert $(\bar{\svec},\bar{a})$ into $B$ with priority $P + \epsilon_p$
%\EndIf
%\State Insert $(\bar{S},\bar{A},\bar{S}',\bar{R}, \bar{\gamma})$ into $B$ with priority $P$
%\State {\bf If} $P >\beta$ Insert $(\bar{S},\bar{A},\bar{S}',\bar{R})$ into PQueue
\EndFor
\EndFor
\EndWhile
\end{algorithmic}
\end{algorithm}

\begin{algorithm}[t]
	\caption{Experience Replay with Predecessors and Prioritization}\label{alg_er}
	\label{alg_erwrbf}
	\begin{algorithmic}
		\State $n$ is the number of planning steps, $\alpha$ is Q-learning stepsize, initial replay buffer $B$ is empty, $b$ is the buffer size limit, $\delta_t$ is the temporal difference error at time step $t$, $\epsilon_p$ is the smoothing parameter to make priority non-zero
\While {true}
	\State Get transition $(\svec, a, \svec', r, \gamma)$
	\State Q-learning update with transition using stepsize $\stepsize$ 
	\State $P \gets | r + \gamma \max_{a'} \hat{q}(\svec',a') - \hat{q}(\svec,a)|$
	\State Append $(\svec_t, a_t, \svec_t', r_t, \gamma_t)$ to $B$ with priority $P + \epsilon_p$
	\State Set priority of $(\svec_{t-1}, a_{t-1}, \svec_{t-1}', r_{t-1}, \gamma_{t-1})$ to $P + \epsilon_p$
	%\If{$|B|>b$} remove the oldest sample in $B$
	%\EndIf
	\For {$n$ times}
	\State $(\svec, a, \svec', r, \gamma) \gets$ sample from $B$ by priority
	\State Q-learning update with stepsize $\frac{\stepsize}{\sqrt{n}}$ 
	\State Update priority of $(\svec, a, \svec', r, \gamma)$
	\EndFor
\EndWhile
	\end{algorithmic}
\end{algorithm}

%\begin{figure}[t]
%	\centering
%	\includegraphics[width=0.5\textwidth]{figures/kdeexplain.pdf}
%	\caption{\small KDE approximation, REM approximation, histograms and true PDF. The figure showed KDE approximation to the true bimodal probability density function with bandwidth $0.15$. The total number of samples is $500$ while our REM approximation only uses $300$ samples. The actual samples are reflected by the short black segments at the bottom. KDE can be intuitively thought as a smooth version of the histogram, avoiding the case of zero density for some samples. This figure is for illustrating purpose only. We first generate samples from the true PDF and then use the two approaches to learn it. We do not do any strict sweep for picking bandwidth.}
%	\label{fig_kderem}
%\end{figure}

% \vspace{0.2cm}
\subsection{Computing Conditional Covariances}

To sample from REMs, as in Algorithm~\ref{alg_sampling_from_REM_forward}, we need to be able to compute the  conditional covariance.
Recall that 
\begin{align*}
\beta_i(\svec, a) &= \tfrac{c_i}{N(\svec, a)} \ k_{\svec}(\svec, \svec_i) k_{a}(a, a_i) \\
&\text{ where } N(\svec, a) = \sum_{i=1}^\ksize c_i \ k_{\svec}(\svec, \svec_i) k_{a}(a, a_i)\\
p(\svec', r, \gamma | \svec, a) &= \sum_{i=1}^\ksize \beta_i(\svec, a)  k_{\svec',r,\gamma}((\svec', r, \gamma), (\svec_i', r_i, \gamma_i))
.
\end{align*}
This is not necessarily the true conditional distribution over $\svec', r, \gamma$, but it is the conditional distribution under our model.
%To compute the covariance for $\svec', r, \gamma$, we average the following conditional covariance across all $\svec, a$.

It is straightforward to sample from this model, using $\beta_i$ as the coefficients, shown in Algorithm \ref{alg_sampling_from_REM_forward}.
The key detail is computing a conditional covariance, described below.

Given sample $\xvec = (s, a)$, the conditional mean is
\begin{align*}
&\muvec(\svec,a) = \int_{(s',r,\gamma)}^{} [\svec', r, \gamma] p(\svec', r, \gamma|\svec,a) d(\svec',r,\gamma) \\
&\!\!\!\!= \int_{(s',r,\gamma)} \!\!\!\!\!\!\!\!\!\!\! [\svec', r, \gamma]  \sum_{i=1}^\ksize \beta_i(\svec, a) k_{\svec', r, \gamma}((\svec', r, \gamma), (\svec_i', r_i, \gamma_i)) d(\svec', r, \gamma)  \\
&\!\!\!=\sum_{i=1}^\ksize \beta_i(\svec, a) \!\!  \int_{(s',r,\gamma)} \!\!\!\!\!\!\!\!\!\!\! [\svec', r, \gamma]  k_{\svec', r, \gamma}((\svec', r, \gamma), (\svec_i', r_i, \gamma_i)) d(\svec', r, \gamma) \\
&= \sum_{i=1}^\ksize \beta_i(\svec, a)  [\svec_i', r_i, \gamma_i]
.
\end{align*}
Similarly, we can compute the conditional covariance
%
% \small
\begin{align}
%\Sigmamat_{\svec', r, \gamma}(\svec, a) % We dont use underscore for mu, so we wont for this either
&\Sigmamat(\svec, a) = \label{eq_conditionalcovariance}\\
&\sum_{i=1}^\ksize \beta_i (\svec,a) \left([\svec_i', r_i, \gamma_i] - \muvec(\svec,a) \right) \left([\svec_i', r_i, \gamma_i] - \muvec(\svec,a) \right)^\top \nonumber
\end{align}
% \normalsize
 %
This conditional covariance matrix more accurately reflects the distribution over $[\svec', r, \gamma]$, given $\svec, a$. A covariance over $[\svec', r, \gamma]$ would be significantly larger than this conditional covariance, since it would reflect the variability across the whole state space, rather than for a given $\svec, a$. For example, in a deterministic domain, the conditional covariance is zero, whereas the covariance of $[\svec', r, \gamma]$ across the space is not. If one consistent covariance is desired for $[\svec', r, \gamma]$, a reasonable choice is to compute a running average of conditional covariances across observed $\svec, a$.  

\subsection{Details on Prototype Selection}\label{app_prototypes}

We use a prototype selection strategy that maximizes $\log \det (\Kmat + \eye)$  \cite{schlegel2017adapting}. There are a number of parameters, but they are intuitive to set and did not require sweeps. The algorithm begins by adding the first $\ksize$ prototypes, to fill up the budget of prototypes. Then, it starts to swap out the least useful prototypes as new transitions are observed. The algorithm adds in new prototypes, if the are sufficiently different from previous prototypes and increase the diversity of the set. The utility increase threshold is set to $0.01$; this threshold simply avoids swapping too frequently, which is computationally expensive, rather than having much impact on quality of the solution. 

A component of this algorithm is a k-means clustering algorithm, to make the update more efficient. We perform k-means clustering using the distance metric $1.0 - k(\xvec, \yvec) = 1.0 - \exp^{-0.5(\xvec - \yvec)^\top \Sigmamat^\inv (\xvec - \yvec)}$, where $\Sigmamat$ is the empirical covariance matrix for transitions $(\svec, a, \svec', r, \gamma)$. The points are clustered into blocks, to speed up the computation of the log-determinant. The clustering is re-run every $10$ swaps, but is efficient to do, since it is started from the previous clustering and only a few iterations needs to be executed. 
%We do ten random restarts multiple ($10$) starts for k-means and pick up the clustering with most even sizes across clusters. 
%When computing the kernel values, we use a constant factor $0.5$ to scale covariance matrix for the sample $(s, a, s', r, \gamma)$ as the bandwidth parameter $\Sigmamat$. 

%\mytodo{Add in any pertinent details on bandwidth selection}
%We do parameter sweep over a better 
%\mytodo{Add in any pertinent details on sampling}

\subsection{Additional Pseudocode for REMs}\label{app_pseudocode}

The pseudocode for the remaining algorithms is included below, in Algorithm 3-6. Note that the implementation for REMs can be made much faster by using KD-trees to find nearest points, but we do not include those details here.

\begin{algorithm}[h]
\caption{Update REM($(\svec,a,\svec',r, \gamma)$)}
\label{alg_remupdate}
\begin{algorithmic}
\State $\triangleright$ Input transition $(\svec,a,\svec',r, \gamma)$
\State Update prototypes with $(\svec,a,\svec',r, \gamma)$ (see Appendix \ref{app_prototypes})
\State $\triangleright$ Update conditional weightings
	\For {$i \in \{1, \ldots, \ksize\}$}
		\State $\rho_i \gets k_{\svec}(\svec, \svec_i) k_a(a, a_i)$
		\State $\rho^r_i \gets k_{\svec}(\svec', \svec'_i) k_a(a, a_i)$
		\State $\cvec_i  \gets (1- \rho_i) \cvec_i + \rho_i k_{\svec', r, \gamma}((\svec', r, \gamma),(\svec'_i, r_i, \gamma_i))$
		\State $\cvec^r_i  \gets (1- \rho^r_i) \cvec^r_i + \rho^r_i k_{\svec}(\svec,\svec_i)$
	\EndFor
\end{algorithmic}
\end{algorithm}

%\begin{figure}[t]
%	\centering
%	\includegraphics[width=8cm]{figures/env_gridworld}
%	\caption{\small Gridworld }
%	\label{fig:gridworld}
%\end{figure}

%\begin{figure}[t]
%	\centering
%	\includegraphics[width=8cm]{figures/env_blockworld}
%	\caption{\small Blockworld }
%	\label{fig:blockworld}
%\end{figure}

%\begin{figure}[t]
%	\centering
%	\includegraphics[width=8cm]{figures/env_riverswim}
%	\caption{\small Riverswim }
%	\label{fig:riverswim}
%\end{figure}

\newcommand{\bfactor}{f}

\begin{algorithm}[h]
\caption{SamplePredecessors($\svec_{t+1}, a_t, \mathcal{M}$)}
\label{alg_sps}
\begin{algorithmic}
\State $\triangleright$ Input model $\mathcal{M}$, state $\svec_{t+1}$, predecessor action $a_t$
\State Set $\bfactor$ = number of predecessors to sample
\State If $p(a_t | \svec_{t+1}) = 0$, return emptyset
%\State where $p(a_t | \svec_{t+1}) = \sum$
%\State Compute $\cvec_a$ for $p(a_t | S_{t+1} = s')$ according to algorithm 3
%\State Compute $k(s', \mathcal{M})$ 
%\State specify $p(a_t | S_{t+1} = s')$ by setting $p(a_t = a' | S_{t+1} = s') = sum(P_a)$ for each $a' \in \Actions$, $P_a = k(s', \mathcal{M}) \circ \cvec_a \circ k(a', \mathcal{M})$
%
\State Return $\bfactor$ predecessors sampled from $p(\svec_t | \svec_{t+1}, a_t)$
\State using sampling Algorithm \ref{alg_sampling_from_REM_reverse}
\end{algorithmic}
\end{algorithm}

\begin{algorithm}[h]
	\caption{Sampling $\svec', r, \gamma$ from REMs, given $\svec, a$}
	\label{alg_sampling_from_REM_forward}
	\begin{algorithmic}
	\State $\triangleright$ Compute coefficients $\beta_i(\svec, a)$
	\State $N(\svec, a) \gets \sum_{i=1}^\ksize c_i \ k_{\svec}(\svec, \svec_i) k_{a}(a, a_i)$
	\For{$i \in \{1, \ldots, \ksize\}$}
		\State $\beta_i  \gets \tfrac{c_i}{N(\svec, a)} \ k_{\svec}(\svec, \svec_i) k_{a}(a, a_i)$
	\EndFor
	\State Sample $j \in  \{1, \ldots, \ksize\}$ accord. to probabilities $\beta_1, \ldots, \beta_\ksize$ 
	\State Compute conditional covariance $\Sigmamat(\svec, a)$ using \eqref{eq_conditionalcovariance}
	\State Return $(\svec', r, \gamma)$ sampled from $\mathcal{N}((\svec'_j, r_j, \gamma_j), \Sigmamat(\svec, a))$
	\end{algorithmic}
\end{algorithm}

\begin{algorithm}[h]
	\caption{Sampling $\svec$ from REMs, given $a, \svec'$}
	\label{alg_sampling_from_REM_reverse}
	\begin{algorithmic}
	\State $\triangleright$ Compute coefficients $\beta_i(\svec, a)$, conditional weights $c^r_i$ for predecessor sampling
	\State $N(\svec, a) \gets \sum_{i=1}^\ksize c^r_i \ k_{\svec}(\svec', \svec'_i) k_{a}(a, a_i)$
	\For{$i \in \{1, \ldots, \ksize\}$}
		\State $\beta_i  \gets \tfrac{c^r_i}{N(\svec, a)} \ k_{\svec}(\svec', \svec'_i) k_{a}(a, a_i)$
	\EndFor
	\State Sample $j \in  \{1, \ldots, \ksize\}$ accord. to probabilities $\beta_1, \ldots, \beta_\ksize$ 
	%\State Compute conditional covariance $\Sigmamat(\svec, a)$ using \eqref{eq_conditionalcovariance}
	\State Return $\svec$ sampled from $\mathcal{N}(\svec_j, \Hmat_s)$
	\end{algorithmic}
\end{algorithm}

\begin{figure*}[t]
	\centering
	\includegraphics[width=0.95\textwidth]{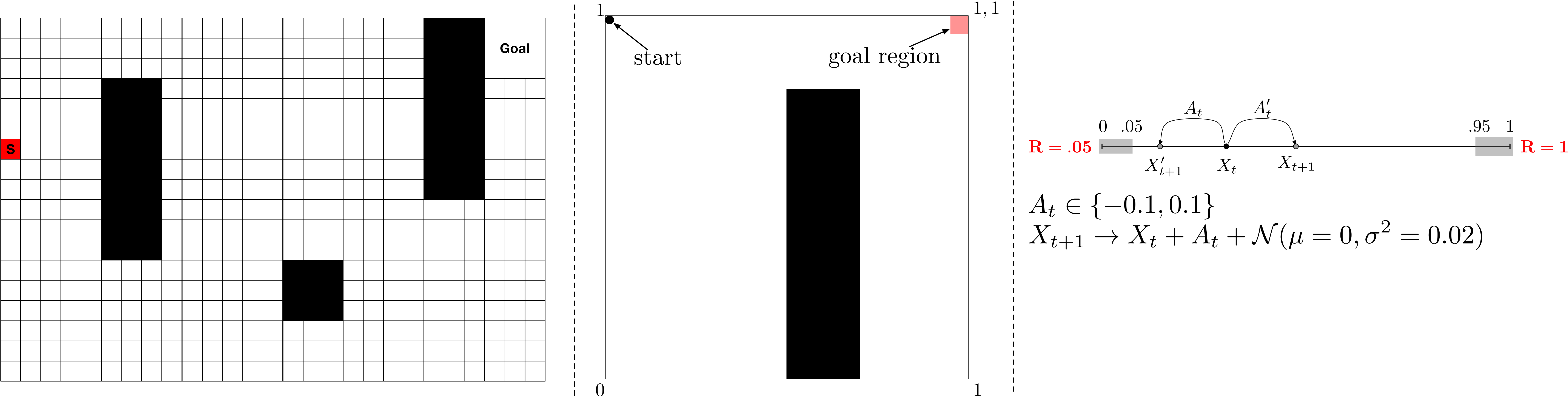} 
	%	\subfigure[Deterministic]{
	%		\includegraphics[width=0.9\textwidth]{figures/tabularDet} \label{fig:grid486 states}} \\
	%		\vspace{-0.25cm}
	%	\subfigure[Stochastic]{
	%		\includegraphics[width=0.9\textwidth]{figures/tabularStoc} \label{fig:grid486 states}} 
	%	\subfigure[864 states, Deterministic]{
	%		\includegraphics[width=\figwidththree]{figures/tabular/states_864/deterministic.png} \label{fig:grid864states}}
	%	\subfigure[1350 states, Deterministic]{
	%		\includegraphics[width=\figwidththree]{figures/tabular/states_1350/deterministic.png}\label{fig:grid1350states}}
	%	\subfigure[486 states, Stochastic]{
	%		\includegraphics[width=\figwidththree]{figures/tabular/states_486/stochastic.png} \label{fig:grid486states_st}} 
	%	\subfigure[864 states, Stochastic]{
	%		\includegraphics[width=\figwidththree]{figures/tabular/states_864/stochastic.png} \label{fig:grid864states_st}}
		%\begin{minipage}{\figwidththree}
	%	\vspace{-10.0cm}
	\caption{ 
		From left to right: Tabular Gridworld, Continuous Gridworld, River Swim}\label{fig:envs} 
	%\end{minipage}
\end{figure*}

\section{Domain Details}
% The domains are depicted in Figure \ref{fig:envs}. We describe each domain in detail below. 
\myparagraph{Tabular Gridworld.}
The tabular grid is a discounted episodic problem with obstacles and one goal, as illustrated in Figure \ref{fig:envs}. The reward is 0 everywhere except the transition into goal, in which case the reward is +100. The discount rate is $\gamma = 0.95$. The agent can take four actions. In the stochastic variant each action takes the agent to the intended next state with probability 0.925, or one of the other three adjacent states with probability 0.025. 

The modelling choices in Dyna are different for the deterministic and stochastic variants. In the deterministic setting, Dyna uses a table to store next state and reward for each state and action; in stochastic, it estimates the probabilities of each observed transition via transition counts.

\myparagraph{Continuous Gridworld.}
The Continuous Gridworld domain is an episodic problem with a long wall as the obstacle between the start state and goal state, where the wall has one hole for the agent to pass through. The state space consists of points $(x, y) \in [0, 1]^2$. The wall in the middle has a width of $0.2$. The hole on the wall is located at $(0.5, 1.0)$ with height $0.2$, hence the hole area is $\{(x, y) : (x, y) \in [0.5, 0.7] \times [0.8, 1.0] \}$. At each step the agent can choose from actions $\{up, down, right, left\}$ to move. The successful move probability is $0.9$; otherwise a random action is taken, with stepsize $0.05 + \mathcal{N}(0, 0.01)$ where $\mathcal{N}(0, 0.01)$ is Gaussian noise with mean $0$ and variance $0.01$. The agent always starts at $(0.0, 1.0)$ and the goal area is $x \ge 0.95, y \ge 0.95$. The reward is $1$ in the goal area, and otherwise is zero. 

\myparagraph{River Swim.}
The continuous River Swim domain is a modified version of the tabular River Swim domain proposed in \cite{strehl2008modeliemdp}. The state space is $[0, 1]$ and the action space is $\{left (-), right (+)\}$. The move step size is $0.1$, changing the state each time by $0.1$ unit. The goal of the agent is to swim upstream (move right) towards the end of the chain, but the dynamics of the environment push the agent downstream, to the beginning of the chain. 
The agent receives a reward of $0.005$ if it is at the beginning of the chain, in region $s \in [0, 0.05]$,
but receives a much larger reward of $1.0$ if it manages to get to the end of the chain, in region $s \in [0.95, 1.0]$.
Otherwise, it receives a reward of zero. 
The task is continuing, with $\gamma = 0.99$.
The left action always succeeds, taking the agent left. 
The right action, however, can fail.
%When the right action is chosen, the agent moves to the right with probability $0.35$, where at the end of the chain it simply stays in that region. Otherwise, the right action fails, and the agent goes left.
Once a right action is chosen, if it is at the beginning of the chain, the agent moves to the right with probability 0.4 and otherwise stays at original position; if it is at the end of the chain, it moves left with probability 0.4 or stays original position with probability 0.6; and if it is at the middle of the chain, the agent moves  right with probability 0.35, left with probability 0.05, or stays with probability 0.6.
% with probability $0.05$ or stays in the same position with probability $0.6$. 
%The agent can move left with probability $0.4$ if it is at the end of the chain otherwise hold. 
Once the move direction is determined, zero-mean Gaussian noise with variance $0.02$ is added to the move amount.  The optimal policy is to always take the right action and a suboptimal policy that agents can get stuck in is to always take the left action. In the experiments section, we present a table showing different percentage of accumulative reward of each algorithm to that of optimal policy. 

%Marthac: these figures are not yet ready, in my opinion, and we have not explained them sufficiently
%Here we attach additional figures to show the distance of each algorithm's accumulative reward to optimal policy's accumulative reward. Please refer to Figure~\ref{fig:fa_exp_sw}. 

%  \begin{figure*}[t]
%  	\vspace{-1cm}
%  	\centering
%  	\subfigure[River Swim, Comparing Search-control]{
%  		\includegraphics[width=\figwidthtwo]{figures/functionapp/sw_compare_dynaER.pdf} \label{fig:swdyna}} 
%  	\subfigure[River Swim, Comparing Models]{
%  		\includegraphics[width=\figwidthtwo]{figures/functionapp/sw_compare_dyna_base.pdf}\label{fig:swbase}} 
%  	%\begin{minipage}{\figwidththree}
%  %		\vspace{-7.0cm}
%  		\caption{ 
%  			Results on RiverSwim showing \textbf{distance of each algorithm's accumulative reward to optimal policy's accumulative reward}. The results are averaged over $100$ runs except neural network model is $30$ runs due to time efficiency consideration. When the curve becomes almost flat, the corresponding agent reaches the optimal policy. Other dyna baselines, such as neural network version or linear dyna, cannot learn the model quickly. There could be an issue if the model learning is slower then the Q learning itself.
%  		}\label{fig:fa_exp_sw}
%  	%\end{minipage}
%  	%\vspace{-0.5cm}	
%  \end{figure*}

\section{Parameter Settings}

All the domains use a tile coding with the number of tilings set to 1 and the number of tiles (discretization) set to 16. For the 2-dimensional Continuous Gridworld, the memory size is set to 2048 and for the 1-dimensional River Swim domain, the memory size is set to 512. 
All algorithms use $10$ planning steps (replay steps). 
For all experience replay algorithms, the buffer size is $1000$ and for all Dyna variants, the search-control queue size is $1000$.
All algorithms use Q-learning, with a sweep over step-sizes in the range $\{0.015625, 0.03125, 0.0625, 0.125, 0.25, 0.5, 1.0\}$ and $\epsilon = 0.1$. 
In the Gridworlds, weights are initialized to zero. In River Swim, the weights are initialized to 1, to provide some optimism and so reduce the number of steps until the right side of the river is first reached. This initialization does not change the conclusions or algorithm ordering, but simply decreased the runtime for initial exploration for all algorithms. 
% Why is RS epsilon = 0

For the learned models, there are additional parameter choices, which we highlight for each model below. 

\myparagraph{REM choices.}
The parameter choices for REM include the number of prototypes and the kernel choices. 
The budget given to the model is $\ksize = 1000$ prototypes.
As mentioned in the text, Gaussian kernels are used. The bandwidth for the kernel over states, $k_\svec$, is set proportional to $1/\ksize$,
and so it was effective to set it to a small number of 0.0001. 
% $\{0.00005, 0.0001, 0.0002\}$ --> this was a pretty narrow range. If possible, it is just better to set it to a small value, and we should run our experiments that way if possible.
The bandwidth for $k_{\svec', r, \gamma}$ is set using conditional covariance. Details on both computing the conditional covariance, and the method to select prototypes are given in Appendix \ref{app_rem}.

\myparagraph{Linear model choices.}
The learned linear model consists of a matrix and vector, $\Fmat_a$ and $\bvec_a$, for each action $a$ such that $\phivec' = \Fmat_a \phivec$ and $r = \bvec_a \phivec$ give the expected next feature vector and reward, from the current state with feature vector $\phivec$ (tile coding) after taking action $a$. 
A separate linear model needs to be learned for the reverse direction, predicting the expected previous feature vector, from the given feature vector if an action $a$ had been taken. These models are learned using a linear regression update on each step, as proposed in the original Linear Dyna algorithm. The stepsize for learning these models is swept over $\{0.03125, 0.0625, 0.125, 0.25\}$. 
To enable a linear model to be learned for each action, the tile coding had to be done separately for each action. The memory size for all the actions is the same, however.
%memory size as it has a separate weight vector for each action. Hence its memory size is the above specified memory size divided by number of actions. 

\myparagraph{Neural network model choices.} As for the linear models, two neural networks are learned, one for the forward model and one for the predecessor model. Both networks have two hidden layers with $40$ and $20$ relu units respectively. In the forward model, the input is the state (observation) and the output is a vector composed of the next state, reward and discount, for each action. The output, therefore, is the dimension of $[S', R, \gamma]$ times the number of actions. The predecessor network is similar, but the output corresponds to a predecessor state. Additionally, because some actions may not lead to the input state, for the predecessor network, we also output probabilities for each action. The activations used for the state, reward, discount and probability are respectively relu, identity, sigmoid and sigmoid. We used the Adam optimizer, with an exponential decay rate of $0.8$ with a decay frequency of $3000$ steps. We swept the initial learning rate in the range $\{0.001, 0.01\}$. Experience replay is used to train the network, using a recency buffer of size $1000$ and a mini-batch size of $32$ for training.

%
%\section{REM Algorithmic details}
%\subsection{Kernel Density Estimation (KDE) and REM}
%KDE method, as a way of learning empirical probability density function, has some advantages over histogram learning. The basic illustration of the method is showed in figure~\ref{fig:kde}. Our REM model, with prototype selection strategy, only requires a subset of samples to cover the space well and additionally learn weights associated with those prototypes. Observing that only conditional probabilities are needed when sampling, we further designed the approach to directly learning the conditional weights $c_i$ as showed in below equation 
%\begin{align*}
%	&p(s',r,\gamma|s,a) \\
%	&=  \alpha \sum_{i=1}^\ksize c_i k_{\svec', r, \gamma}((\svec', r, \gamma), (\svec_i', r_i, \gamma_i))  k_{\svec} (\svec_i, \svec) k_a (a_i, a)
%\end{align*}
%where $\alpha$ is some normalization constant to make sure the integration of PDF equals to 1. The details are explained in section $4.2$.

%\section{Algorithmic details}
%\subsection{Kernel Density Estimation (KDE) overview}
%KDE method, as a way of learning empirical probability density function, has some advantages over histogram learning. The basic illustration of the method is showed in figure~\ref{fig:kde}. Our REM model, with prototype selection strategy, only requires a subset of samples to cover the space well and additionally learn weights associated with those prototypes. 

\end{document}